\documentclass{article}

\usepackage[preprint]{spconfa4}
\usepackage{amssymb,amsmath}
\usepackage{framed,multirow}
\usepackage{latexsym}
\usepackage{array}
\usepackage{balance}
\usepackage{booktabs}
\usepackage{caption}
\usepackage{color}
\usepackage{graphicx}
\usepackage{multirow}
\usepackage{times}
\usepackage{url}
\usepackage{threeparttable}
\usepackage{float}
\usepackage{CJK}
\usepackage{color,xcolor}
\usepackage{longtable}

\newcommand\etal{\emph{et~al.}}
\newcommand\ie{\emph{i.e.}}
\newcommand\etc{\emph{etc.}}

\newcommand{\p}[1]{\textcolor{black}{#1}}
\newcommand{\REO}[1]{\textcolor{black}{#1}}

\newcommand{\ZJH}[1]{\textcolor{black}{#1}}

\begin{document}
\UseRawInputEncoding

\title{A Survey of Human-in-the-loop for Machine Learning}

\name{Xingjiao Wu$^{1,2}$, Luwei Xiao$^{2}$, Yixuan Sun$^{3}$, Junhang Zhang$^{2}$, Tianlong Ma$^{1,2}$\thanks{$^*$Corresponding author.}, Liang He$^{1,2,*}$}
\address{\small{1 Shanghai Key Laboratory of Multidimensional Information Processing, East China Normal University, Shanghai, China}\\
\small{2 School of Computer Science and Technology, East China Normal University, Shanghai, China}\\
\small{3 Fudan University, Shanghai, China}\\
{\small \{wuxingjiao2885, louisshaw008, madmaxkgb, junhangzhang68\}@gmail.com, \{tlma, lhe\}@cs.ecnu.edu.cn}
}

\maketitle

\begin{abstract}
\REO{Machine learning has become the state-of-the-art technique for many tasks including computer vision, natural language processing, speech processing tasks, \ZJH{\etc}
However, the unique challenges posed by machine learning suggest that incorporating user knowledge into the system can be beneficial.
The purpose of integrating human domain knowledge is also to promote the automation of machine learning.
Human-in-the-loop is an area that we see as increasingly important in future research due to the knowledge learned by machine learning cannot win human domain knowledge. }
Human-in-the-loop aims to train an accurate prediction model with minimum cost by integrating human knowledge and experience.
Humans can provide training data for machine learning applications and directly accomplish tasks that are hard for computers in the pipeline with the help of machine-based approaches.
In this paper, we survey existing works on human-in-the-loop from a data perspective and classify them into three categories with a progressive relationship: (1) the work of improving model performance from data processing, (2) the work of improving model performance through interventional model training, and (3) the design of the system independent human-in-the-loop. Using the above categorization, we summarize \ZJH{the} major approaches in the field; along with their technical strengths/ weaknesses, we have a simple classification and discussion in natural language processing, computer vision, and others. Besides, we provide some open challenges and opportunities. This survey intends to provide a high-level summarization for \REO{human-in-the-loop} and \ZJH{ to motivate} interested readers to consider approaches for designing effective \REO{human-in-the-loop} solutions.

\end{abstract}

\begin{keywords}
Human-in-the-loop, machine learning, deep learning.
\end{keywords}

\section{Introduction}
\label{sec:Intro}
Deep learning is a frontier for artificial intelligence, aiming to be closer to its primary goal\ZJH{��}artificial intelligence.
Deep learning has seen great success in a wide variety of applications, such as natural language processing, speech recognition, medical applications, computer vision, and intelligent
transportation system~\cite{dong2021survey}.
The great success of deep learning is due to the larger models ~\cite{brutzkus2019larger}.
The scale of these models has included hundreds of millions of parameters.
These hundreds of millions of parameters allow the model to have more degrees of freedom enough to awe-inspiring description capability.

\begin{figure}
    \centering
    \includegraphics[width=0.80\linewidth]{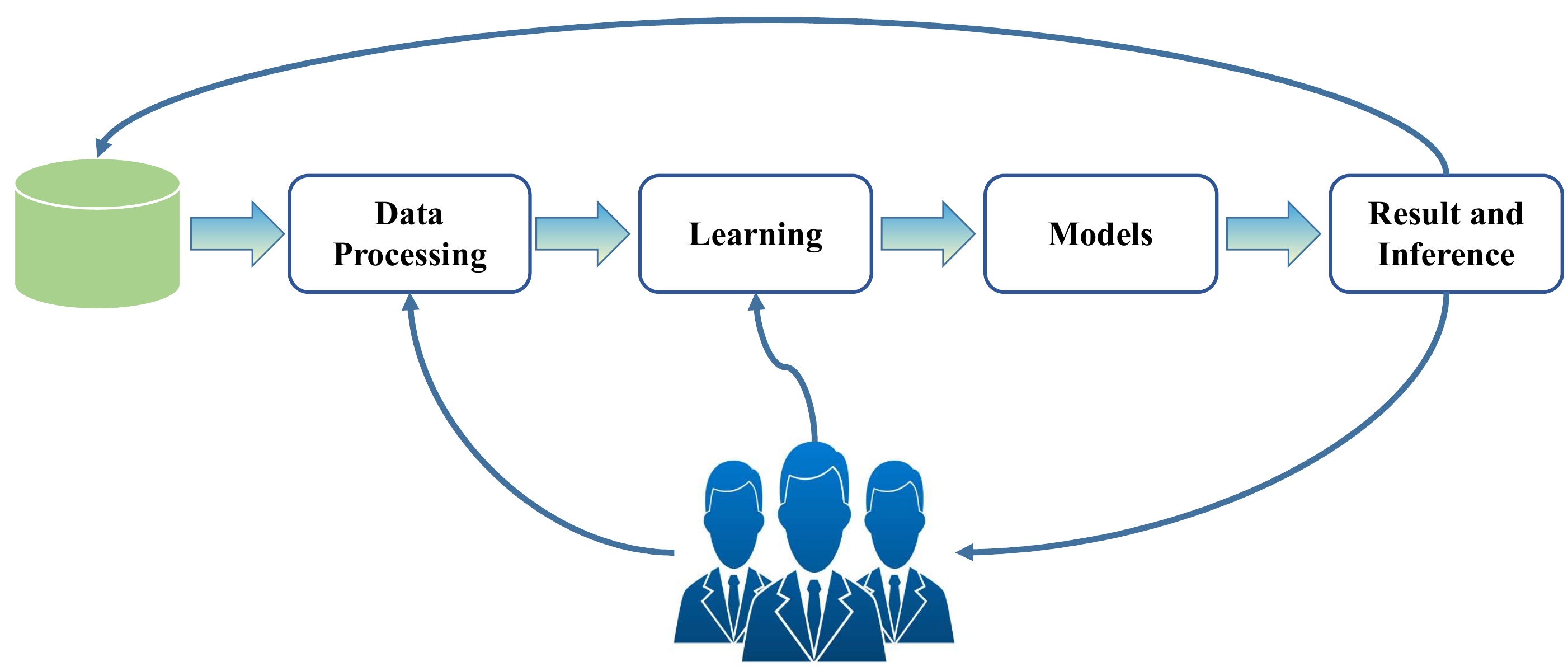}
    \caption{The development cycle of model.
    }
    \label{fig1}
\end{figure}

However, the large number of parameters requires a massive amount of training data with labels~\cite{zhou2014learning}.
Improving model performance by data annotation has two crucial challenges.
On the one hand, the data growth rate is far behind the growth rate of model parameters, \ZJH{which} has primarily hindered the further development of the model.
On the other hand, the emergence of new tasks has far exceeded the speed of data updates, and annotating all samples is \p{labor-intensive and time-consuming.}
To tackle this challenge, many researchers build new datasets by generating samples, thereby speeding up model iteration and reducing the cost of data
annotation~\cite{li2020layoutgan,zhao2020differentiable,shen2021heterogeneous}.
Besides, \p{a group of researchers devise pre-trained models and transfer learning to solve this challenge~\cite{qiu2020pre, zaib2020short,bahrami2021joint}, such as Transformers~\cite{ vaswani2017attention}, BERT~\cite{devlin2018bert} and GPT~\cite{radford2018improving}.}
These works have achieved incredible results.

\ZJH{Unfortunately}, the generated data is only used as base data to initialize the model.
To obtain a high-precision usable model, \p{labeling and updating specific data is often necessary.}
So various work based on weak supervision has been proposed~\cite{habermann2020deepcap, wang2020weak}.
A great many researchers have proposed using few-shot to push the model to learn from fewer samples~\cite{ jia2021survey}.

\subsection{Significance of \REO{Human-in-the-loop}}

Integrated a priori knowledge in the learning framework is an effective means to deal with sparse data, as the learner does not need to induce the knowledge from the data~\cite{diligenti2017integrating}.
\p{More recently, an ever-growing number of researchers have made efforts to incorporate pre-training knowledge into their learning framework}~\cite{chen2020deep, lin2020deep, hartmann2019deep}.
As special agents, humans have rich prior knowledge.
\REO{If the developer \ZJH{can} encourage the machine to engage with learning human wisdom and knowledge, it would help deal with sparse data, especially in medical fields like clinical diagnosis and lack of training data~\cite{zhang2018towards, zhang2019leveraging, holzinger2019interactive, zhuang2017challenges}.
Furthermore, recent advances in cognitive science and human-machine interaction have suggested that human-related elements, such as emotional state and practical capability, impact human teaching performance and machine learning results on different tasks.}

\p{A multitude of researchers have proposed utilizing a conception called} ``\REO{human-in-the-loop}" to tackle the above challenges, mainly addressing these issues by incorporating human knowledge into the modeling process~\cite{kumar2019didn}. Human-in-the-loop conception is an extensive area of research that covers the intersection of computer science, cognitive science, and psychology.

\begin{figure}[t]
    \centering
    \includegraphics[width=0.88\linewidth]{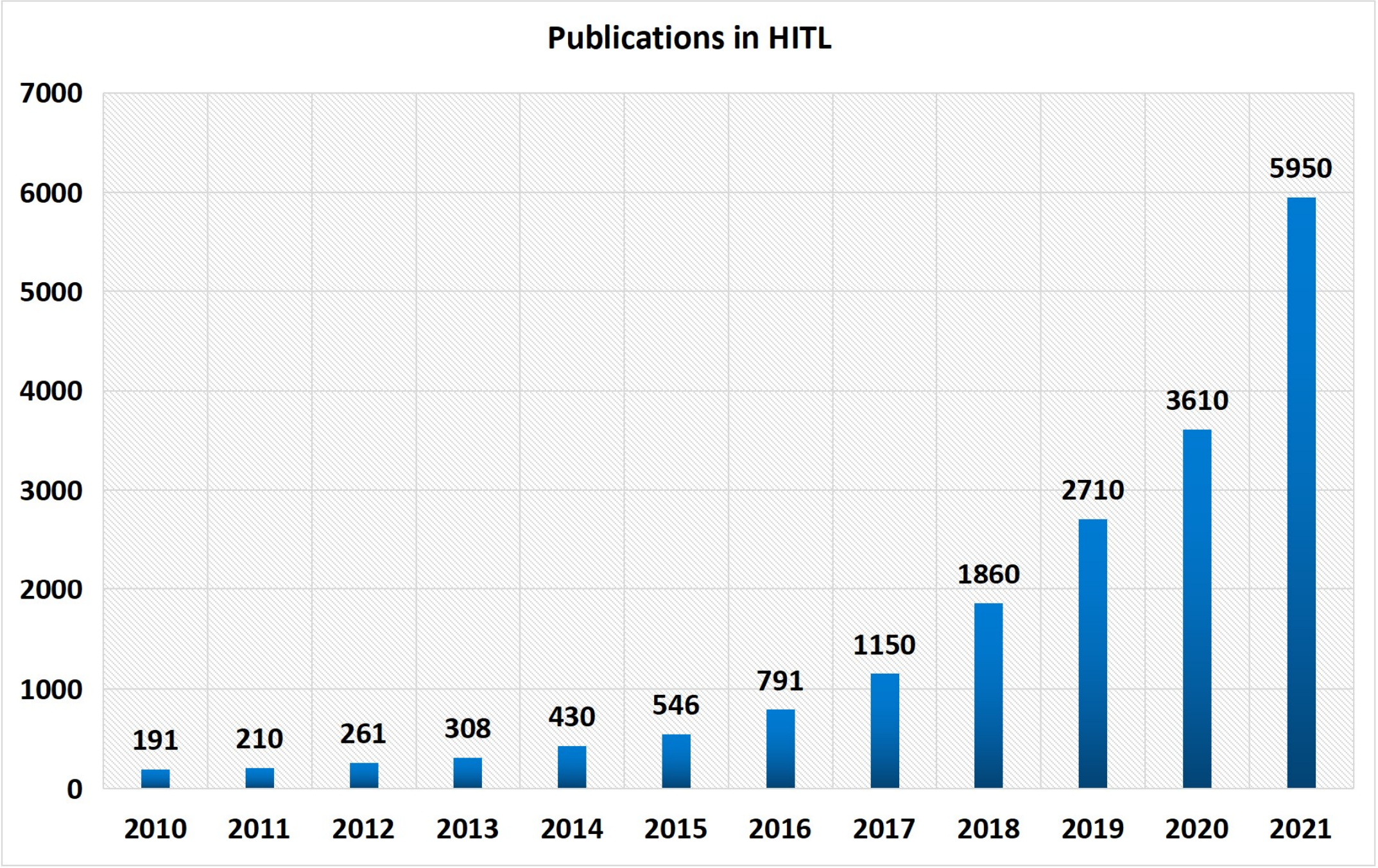}
    \caption{The increasing research interest in the \REO{human-in-the-loop}, obtained through Google scholar search with key-words: ``\REO{human-in-the-loop}"  and ``machine learning".
    }
    \label{fig2}
\end{figure}

As illustrated in Fig.~\ref{fig2}, \REO{human-in-the-loop} (namely ``\REO{human-in-the-loop}" and ``machine learning") is an active research topic in machine learning, and there has been a rich publication in the past ten years.

As shown in Fig.~\ref{fig1}, a conventional machine-learning algorithm typically consists of three parts~\cite{xin2018accelerating}.
The first is data preprocessing, the second is data modeling, and the last is the developer modifying the existing process to improve performance.
Generally speaking, the performance and results of machine learning models are unpredictable, which leads to a large degree of uncertainty in which part of the machine-human interaction is capable of bringing the best learning effect.
Different researchers focus on manual intervention in distinguishable parts.
In this paper, we investigate existing studies on \REO{human-in-the-loop} technology via various implementations of \REO{human-in-the-loop} from different practical perspectives\ZJH{(e.g. Data Processing, Model Training and Inference, and System construction and Application).}
It is crucial to explore how interaction type interplays with other components of a \REO{human-in-the-loop} pipeline to affect the intelligent systems' learning outcomes.
In addition, more research focuses on the design of independent systems to help complete the improvement of the model.
In this paper, we first discuss the work of improving model performance from data processing.
Next, we discuss the work of improving model performance through interventional model training.
Finally, we discuss the configuration of the system independent ``\REO{human-in-the-loop}".

\REO{
We \ZJH{explore} the following questions around human-in-the-loop for machine learning:
\begin{itemize}
\setlength{\itemsep}{0pt}
\setlength{\parsep}{0pt}
\setlength{\parskip}{0pt}
\item What is the human-in-the-loop for machine learning challenges, and what are the possible solutions to advance this research?
\item From a data perspective, what is the research status of human-in-the-loop for machine learning, research challenges, and what are the future directions?
\item From the perspective of model training, what is the research status of human-in-the-loop for machine learning, research challenges, and what are the future directions?
\item From an application perspective, what is the research status of human-in-the-loop for machine learning, research challenges, and what are the future directions?
\end{itemize}
}

\subsection{Importance of this survey}

This paper is a systematic summary and analysis of the research area of \REO{human-in-the-loop} with a focus on the following essential aspects:

\begin{itemize}
\setlength{\itemsep}{0pt}
\setlength{\parsep}{0pt}
\setlength{\parskip}{0pt}
	\item  We \ZJH{had} a comprehensive summary of the work of \REO{human-in-the-loop} so far, \ZJH{and divided} these papers into CV, NLP and other application areas. \ZJH{We connected} these papers in series according to data preprocessing, data annotation, and model training and inference from the perspective of data flow; and finally, we focused on the application based on \REO{human-in-the-loop};
	\item We have classified and compared various methods of \REO{human-in-the-loop}. Through classification and comparison, we have summarized the challenges currently encountered by \REO{human-in-the-loop} and put forward some discussions on solving these challenges;
	\item We also conducted qualitative evaluations and comparisons between different methods to evaluate them consistently. This would assist readers in deciding which method is appropriate for the problem at hand;
    \item  We have also identified a series of important milestones achieved by methods in this area;
	\item In addition to the methodological summary and analysis of \REO{human-in-the-loop}'s work, we also discussed the system construction and application of \REO{human-in-the-loop}. We analyze the \REO{human-in-the-loop} on system construction divided by system components and applications following the baseline of engineering requirements.
\end{itemize}

\REO{Before our work, there were also quiet a few excellent works reviewing human-in-the-loop~\cite{budd2021survey,jung2019human,agnisarman2019survey,benedikt2020human,chai2020human,tehrani2019review}, but we are the first review of human-in-the-loop for machine learning. Much of the previous work is from the perspective of hardware and robotics, \ZJH{while} we are talking about machine learning. In our research process, we also follow on previous experiences and ideas.}
\ZJH{Certainly}, a single article may not cover all methods in this growing field. Nevertheless, we have sought to make this survey as comprehensive as possible.
To help with this goal, we researched and analyzed a good many related references and documents. In addition, we also tried a variety of classification methods to divide the article structure, and finally, we chose to connect the whole work from the perspective of data.

\subsection{Organization of contents}

\REO{Since there has been no similar review before, we explore all papers (since 1990) containing the keywords ``human-in-the-loop" and ``machine learning" by Google Scholar.
Firstly, we screen out relevant articles by reading the titles and abstracts. Secondly, we briefly classify the screened articles. Moreover, we read the screen articles in detail and constantly revised the classification. Finally, we summarize and form the final content according to the classification.}

In Section~\ref{sec:DP}, we investigate the data processing method based on \REO{human-in-the-loop} and then discuss data preprocessing, data annotation, and iterative labeling.
In Section~\ref{sec:MTI}, we summarize and analyze research on model training and reasoning based on \REO{human-in-the-loop}, and discuss \REO{human-in-the-loop} from natural language processing and computer visual perspectives, respectively.
In Section~\ref{sec:SCA}, we review the \REO{human-in-the-loop} on system construction divided by system components and applications and discuss \REO{human-in-the-loop} from the software and hardware integrated perspective, respectively.
In Section~\ref{sec:DF}, we propose a list of challenges based on the results of the survey.
Finally, we conclude our work in Section \ref{sec:conclusion}.

\section{Data Processing}
\label{sec:DP}
At present, deep learning has played an irreplaceable role in many fields~\cite{dong2021survey, khan2021deep}.
The great success of deep learning is due to larger-scale models, which include hundreds of millions of parameters ~\cite{brutzkus2019larger}.
Such a large amount of parameters empower the model with enough degrees of freedom to obtain awe-inspiring description capability. A massive amount of training data with labels are required to deal with a considerable number of parameters~\cite{zhou2014learning}.
However, making annotations requiring much labor is likely to lag behind the growth in model capacity, and available datasets are quickly becoming outdated in size and density~\cite{yu2015lsun}.
\p{So the methods of utilizing unlabeled data to improve the model capability have increasingly gained much attention} ~\cite{zhou2014learning,simeoni2021rethinking,wang2021trust,shi2021boosting}.
The most significant difficulty is based on the fact that unlabeled data usually include incorrect samples, such as disturbing images, defective statements, and violations of constraints. Suppose these inaccurate samples are exactly sampled as the key one, the errors brought about will be fatal~\cite{ren2020not}.
To tackle this challenge, numerous researchers focus on exploring the way to generate a more rich sample space~\cite{li2020layoutgan,niu2020defect,zhao2020differentiable}, trying to develop a universal model such as Transformers~\cite{khan2021transformers}, BERT~\cite{devlin2018bert} or GPT~\cite{radford2018improving} so that the model can learn features more effectively.
Based on these successful methods, researchers then consider the further step: \p{adopting little data to obtain more satisfactory results. So these models are employed for more tasks by fine-tuning and performing incredible results} ~\cite{pham2021classification,chen2020recall,wang2018interactive}.
Although these methods still need to annotate a lot of data, which brings unnecessary trouble, we still noticed that interference in the model performance only some critical samples in the new datasets.
Here goes to a critical issue that needs to be solved urgently,
\emph{How do we find out the key samples, and can we annotate key samples more easily?}

\begin{table*}[t]
\centering
\caption{A overview of representative works in data processing.
DP: data preprocessing; DA: data Annotation; IL: iterative labeling;
CV: Computer Vision;
NLP: Natural Language Processing;
SP: Speech Processing.
}
\label{DPtable1}
\scalebox{0.78}{\begin{tabular}{@{}p{82px}|p{18px}<{\centering}p{18px}<{\centering}p{18px}<{\centering}|p{26px}<{\centering}|p{12px}<{\centering}p{12px}<{\centering}p{12px}<{\centering}p{18px}|p{128px}|c@{}}
\toprule
\multicolumn{1}{c|}{\multirow{2}{*}{Work}} & \multicolumn{3}{c}{Data Processing} & \multicolumn{1}{|c|}{\multirow{2}{*}{Year}} & \multicolumn{4}{c}{Area} & \multicolumn{1}{|c}{\multirow{2}{*}{Task}}  & \multicolumn{1}{|c}{\multirow{2}{*}{\REO{Quantitative results}}}  \\ \cmidrule(lr){2-4} \cmidrule(lr){6-9}
\multicolumn{1}{c|}{}                      & DP      & DA     & IL     & \multicolumn{1}{c|}{}                      & CV  & NLP  & SP  & Other & \multicolumn{1}{c|}{}  & \multicolumn{1}{c}{}                         \\ \cmidrule(r){1-1} \cmidrule(lr){1-11}
Yu~\etal~\cite{yu2015lsun}                                        &         &        & \checkmark      & 2015                                      & \checkmark   &      &     &       & Scene and Object Categories     &\REO{$0.82->0.88(mAP)$}              \\
He~\etal~\cite{he2016human}                                        &         &        & \checkmark      & 2016                                      &     & \checkmark    &     &       & CCG Parser                 &\REO{$84.2\%->85.9\%(F_1)$}                   \\
Self~\etal~\cite{self2016bridging}                                      & \checkmark       &        &        & 2016                                      &    &     &    & \checkmark     & Data Analysis             &\REO{$--$}                    \\
Zhuang~\etal~\cite{zhuang2017hike}                                   & \checkmark       &        &        & 2017                                      &     &      &     & \checkmark     & Knowledge Bases Integration        &\REO{$84.3\%->86.6\%(F_1)$}           \\
Li~\etal~\cite{li2017human}                                        & \checkmark       &        &        & 2017                                      &     &      &     & \checkmark     & Data Integration                  &\REO{$--$}            \\
KIM~\etal~\cite{kim2018human}                                        &         & \checkmark      &        & 2018                                      &     &      & \checkmark   &       & Finding Sound Events           &\REO{$--$}               \\
Doan~\etal~\cite{doan2018human}                                      & \checkmark      &        &        & 2018                                      &     &      &     & \checkmark     & Data Analysis         &\REO{$--$}                         \\
Dong~\etal~\cite{dong2018data}                                       & \checkmark       &        &        & 2018                                      &     & \checkmark    &     & \checkmark     & \small{Data Fusion;DOM extraction}  &\REO{$--$}    \\
Gentile~\etal~\cite{gentile2019explore}                                   & \checkmark       & \checkmark      &        & 2019                                      &     & \checkmark    &     &       & Dictionary Expansion        &\REO{$41.17\% \uparrow$}                  \\
Zhang~\etal~\cite{zhang2019invest}                                    &         & \checkmark      & \checkmark      & 2019                                      &     & \checkmark    &     &       & Entity Extraction          &\REO{$0.8376->0.8644(F_1)$}                    \\
Laure~\etal~\cite{berti2019reinforcement}                                      & \checkmark       &        &        & 2019                                      & \checkmark   & \checkmark    & \checkmark   & \checkmark     & Data Preparation      &\REO{$--$}                        \\
Gurajada~\etal~\cite{gurajada2019learning}                                   & \checkmark       &        &        & 2019                                      &     & \checkmark    &     &       & Entity Resolution                  &\REO{$--$}           \\
Lou~\etal~\cite{lou2019knowledge}                                        & \checkmark       &        &        & 2019                                      &     & \checkmark    &     &       & \small{Knowledge Graph Programming}     &\REO{$98.78\% (top-10~hit)$}               \\
Liu~\etal~\cite{liu2019deep}                                        &         & \checkmark      & \checkmark      & 2019                                      & \checkmark   &      &     &       & Person Re-Identification        &\REO{$45.55\%->71.52\%(mAP)$}              \\
Wallace~\etal~\cite{wallace2019trick}                                    &         & \checkmark      &        & 2019                                      &     & \checkmark    &     &       & Question Answering                  &\REO{$--$}          \\
Fan~\etal~\cite{fan2019interactive}                                        &         &        & \checkmark      & 2019                                      & \checkmark   &      &     & \checkmark     & Network Anomaly Detection       &\REO{$84.71\%(FPR)$}              \\
Krokos~\etal~\cite{krokos2019enhancing}                                     &         &        & \checkmark      & 2019                                      & \checkmark   &      &     & \checkmark     & Knowledge Discovery             &\REO{$67.7\%->88.2\%(Acc)$}              \\
Klie~\etal~\cite{klie2020zero}                                       &         & \checkmark      &        & 2020                                      &     & \checkmark    &     &       & Entity Linking                       &\REO{$35\% \uparrow ( annotation~speed)$}         \\
Chai~\etal~\cite{chai2020human}                                       & \checkmark       &        &        & 2020                                      &     &      &     & \checkmark     & Outlier Detection               &\REO{$88\%->94\%(recall)$}              \\
Butler~\etal~\cite{butler2020human}                                     &         & \checkmark      &        & 2020                                      & \checkmark   &      &     &       & Facial Expressions                 &\REO{$--$}            \\
Ristoski~\etal~\cite{ristoski2020large}                                   & \checkmark       &        &        & 2020                                      &     & \checkmark    &     &       & Relation Extraction                  &\REO{$0.02->0.0321(F_1)$}         \\
Qian~\etal~\cite{qian2020partner}                                       & \checkmark       &        &        & 2020                                      &     & \checkmark    &     &       & Entity Name Understanding           &\REO{$--$}           \\
Le~\etal~\cite{le2020toward}                                         &         & \checkmark      & \checkmark      & 2020                                      & \checkmark   &      &     &       & \small{Self-Annotation For Video Object}  &\REO{$34.1\%->56.6\%(mIoU)$} \\
Bartolo~\etal~\cite{bartolo2020beat}                                    &         & \checkmark      &        & 2020                                      &     & \checkmark    &     &       & Reading Comprehension            &\REO{$39.1\%(F_1)$}             \\
Cutler~\etal~\cite{muthuraman2021data}                                     & \checkmark       &        &        & 2021                                      &     & \checkmark    &     &       & Entity Recognition                   &\REO{$--$}         \\
Meng~\etal~\cite{meng2021towards}                                       &         & \checkmark      &        & 2021                                      & \checkmark   &      &     &       & \small{3D Point Cloud Object Detection}        &\REO{$--$}       \\
Zhang~\etal~\cite{zhang2021generating}                                      &         & \checkmark      &        & 2021                                      & \checkmark   &      &     &       & \small{Screentone and Manga Processing}        &\REO{$--$}       \\
Adhikari~\etal~\cite{adhikari2021iterative}                                   &         & \checkmark      &        & 2021                                      & \checkmark   &      &     &       & Object Detection                        &\REO{$--$}      \\ \bottomrule
\end{tabular}}
\end{table*}

The intuitive idea to solve this concern with a specific method is in three steps:
Select a few samples which models can not recognize.
(1) Use the particular approach to annotate selected samples.
(2) Push the model to learn features from the latest annotated samples.
(3) This idea allows the model to make the most of the data information at the least cost.

\begin{figure}[t]
    \centering
    \includegraphics[width=\linewidth]{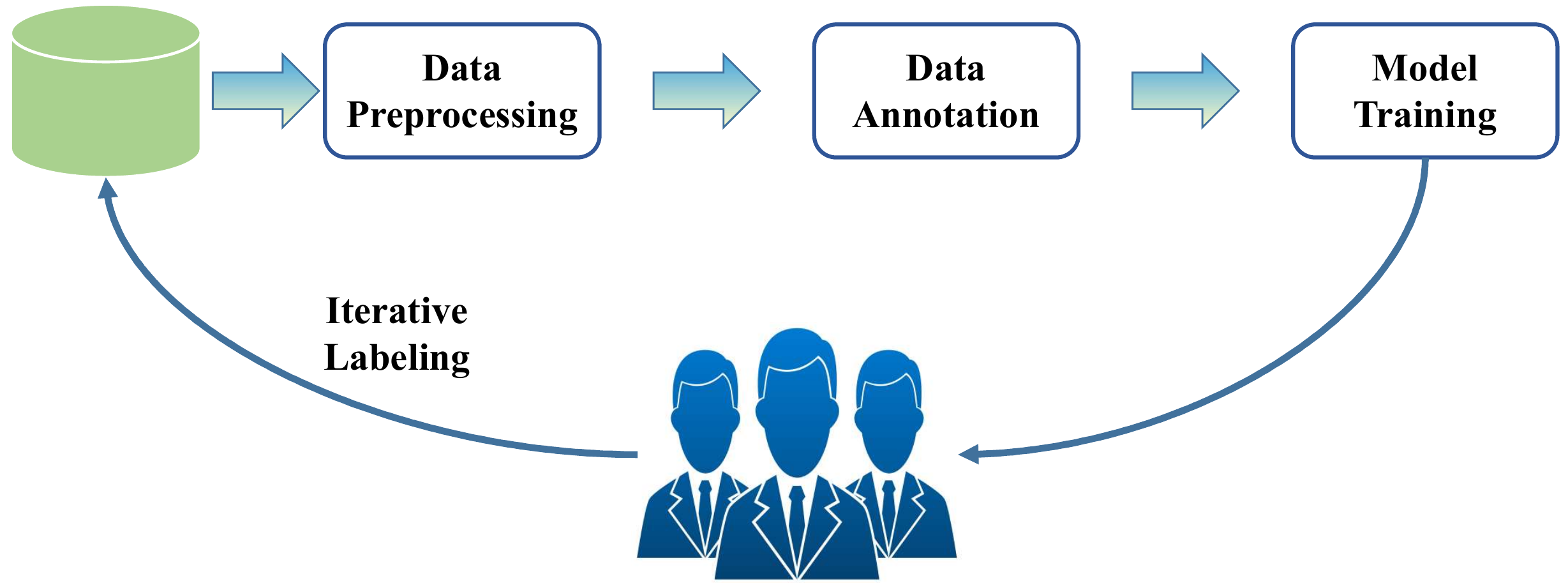}
    \caption{A \REO{human-in-the-loop} data processing pipeline.
    }
    \label{Dp1}
\end{figure}

Multiple researchers try to apply \REO{human-in-the-loop}-based methods to optimize models from the perspective of data. According to surveys, scientists spend about 80\% of their time on data processing compared to model building~\cite{chai2020human}.
We investigated the data processing methods based on \REO{human-in-the-loop} and established a pipeline as shown in Fig.~\ref{Dp1}. We reviewed the representative works in data processing and showed the classification result as shown in Table~\ref{DPtable1}.
This section explores the strengths and deficiencies of data processing with \REO{human-in-the-loop} by demonstrating data preprocessing, data annotation, and iterative labeling.

\subsection{Data Preprocessing}
It is well-known that deep learning is a process of modeling data.
The success of deep learning largely depends on the quality of data, and data analysis plays an irreplaceable role in building a more effective model.
However, it is not easy to find a static method for data analysis, which means data scientists must analyze existing data by employing experts' experience.
The greatest challenge in data analysis lies in the complexity of high-dimensional data, which makes it hard for models to discover data structure. \p{Furthermore, adjusting these parameters dramatically depends on the knowledge of data experts or domain experts.}
Motivated by this phenomenon, Self~\etal~\cite{self2016bridging}~proposed a human-model interactive parameter adjustment mode to facilitate user participation by bridging the gaps between a user's intention and the parameters of a weighted multidimensional scaling model.
Therefore, Doan~\etal~thought establishing a benchmark is an effective means to solve this challenge~\cite{doan2018human}.
Besides, data analysis inevitably involves two considerations: how to carry out automated parameter analysis methods, and the other is how to explore the ability to establish a specific benchmark.
Considering these two issues simultaneously, Laure~\cite{berti2019reinforcement} expanded based on Learn2Clean. They developed automated machine learning approaches (AutoML) that can optimize the hyper-parameters of a considered ML model with a list of by-default preprocessing methods.
This method is devoted to proposing a principled and adaptive data preparation approach to help and learn from the user to select the optimal sequence of data preparation tasks.
With the rapid development of research, researchers are no longer satisfied with solving specific problems in \REO{human-in-the-loop} data analysis (HILDA). They are more concerned with several ``big picture" questions regarding HILDA. Current data analysis technology can correctly obtain the necessary information and knowledge by constructing a knowledge base or graph. However, for the HILDA community or tools, the degree of attention is not enough. Researchers should pay more attention to such issues and make them more popular in the user community to develop data repositories and tools.

\begin{figure}
    \centering
    \includegraphics[width=0.88\linewidth]{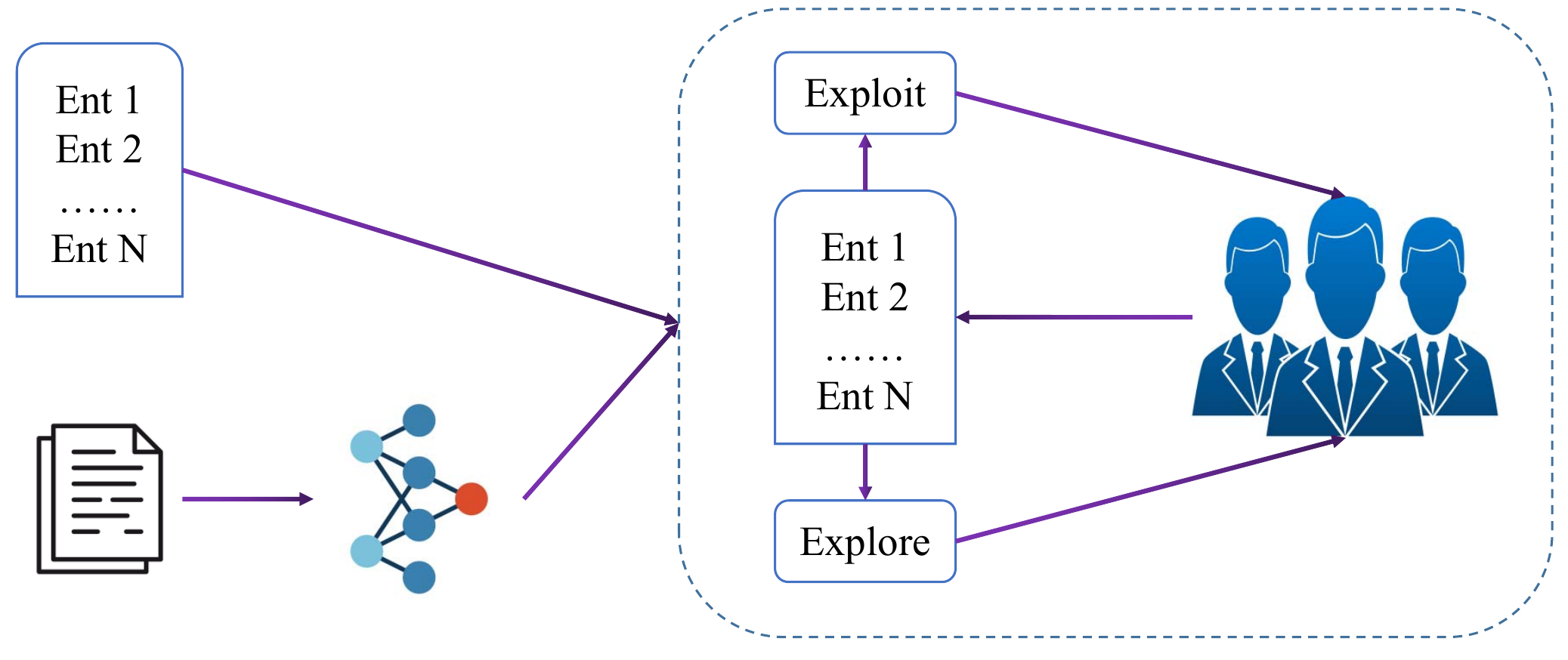}
    \caption{The architecture of explore and exploit.
    }
    \label{Dp2}
\end{figure}

Utilizing \REO{human-in-the-loop} methods to deal with natural language data has inherent advantages over other kinds of data (\ie,~speech recognition, medical applications, computer vision, and intelligent transportation system).
Most of the \REO{human-in-the-loop} methods are used in the information extraction stage.
As Fig.~\ref{Dp2}, Gentile~\etal~\cite{gentile2019explore} propose an interactive dictionary expansion tool using two neural language models.
Many researchers add humans to NLP tasks(such as entity analysis, knowledge graphs, etc.) by using crowdsourcing~\cite{gentile2019explore, berti2019reinforcement, lou2019knowledge}.
Ristoski~\etal~\cite{ristoski2020large}~introduced a method of extracting instances from various web resources, which dramatically improves the performance of the system by introducing human-recycling components.
Besides, this method can integrate the human experience and knowledge to empower machines' accurate intelligence.
Consistent with the method of dictionary expansion proposed before, the core idea of this method is also realized by expanding the existing dictionary.
In addition to the direct annotation, since there are semantic disambiguation phenomena in a few NLP tasks, exploiting the model alone cannot accomplish these tasks. However, the performance of these tasks can be significantly improved by adopting the human knowledge intervention accumulated in unconscious learning.
Qian~\etal~\cite{qian2020partner}~ put forward a deep learning-based entity name understanding system called PARTNER, which provided a more reasonable way of interaction. PARTNER is based on active learning and a weak supervision method.
It is necessary to apply data screening technology for sample selection to find those error-prone samples in the \REO{human-in-the-loop} process.
Cutler~\etal~\cite{muthuraman2021data} presented a method that marked potentially incorrect labels with high sensitivity in the named entity recognition corpus.

Our summary of the previous papers finds that most of the existing \REO{human-in-the-loop} preprocessing studies concern extracting and analyzing complex information in the real world. Nevertheless, there is still very little work that utilizes \REO{human-in-the-loop} technology to perform data preprocessing on CV tasks. \p{We conjuncture that the nature of this phenomenon is that there is a lack of a perfect mode to integrate the human experience into image processing, detailed discussion can be found in Section~\ref{sec:DF}.}

\subsection{Data Annotation}
\label{subsec:Data Annotation}

For new tasks, annotating data is a complex but crucial task to realize artificial intelligence. \p{A considerable number of researchers have proposed employing a \REO{human-in-the-loop}-based method for fast and precise (compared to fragile labeling) operations, especially in NLP and CV domains.}

In NLP tasks, data annotation is bifurcated into two categories. One is the annotation of specific task datasets, such as entity extraction~\cite{gentile2019explore, zhang2019invest}, entity linking~\cite{klie2020zero}, and the other is more abstract tasks, such as Q\&A tasks~\cite{wallace2019trick} and reading comprehension tasks~\cite{bartolo2020beat}.

The entity processing task is critical in NLP, and its success or failure directly affects the performance of NLP~\cite{martinez2020information}.
Currently, there are two main methods for entity extraction, one is to formulate regular expressions for automatic extraction, and the other is the entity mentioned in the manual tagged document.
However, neither of these two strategies can extract entities efficiently and accurately. Zhang~\etal~\cite{zhang2019invest} devised a \REO{human-in-the-loop}-based entity extraction method to obtain the best return on investment in a limited time.
With the deepening of research, we need to handle a growing base of tasks. The emergence of new schemes is beyond our expectations.
Regular expressions can help handle common data, but there is no expected magic for new data never seen before. To mitigate this issue, a few studies proposed approaches to solve the cross-domain problem in entity links.
They find the entities mentioned in the text and filter and discriminate them according to entities sorting information. This method is especially suitable for semantic disambiguation tasks~\cite{klie2020zero}.

How to cope with more complicated tasks is also a focus of research now. Researchers attempt to integrate human experience and knowledge to endow machines with more intelligence.
More specifically, to what extent do neural network models understand the natural language, and can they be further enhanced? To explain this problem and explore more interpretability of the neural network model.
Wallace~\etal~\cite{wallace2019trick} developed an open application system that contains an interactive interface to talk with the machine, thereby generating more Q\&A language materials to collect more research data and help the researcher explain the model predictions.
Bartolo~\etal~\cite{bartolo2020beat} tried three different sets of annotation methods in the reading comprehension task to build a gradually more robust model in the annotation cycle. Significantly, they created a challenging dataset by collecting 36,000 samples.
However, with the enhancement of the cyclic model, the performance gradually deteriorates. In contrast, the more robust model can still learn from the data intensively collected by the weaker model in the loop~\cite{ye2016face}.

As for CV, \REO{human-in-the-loop} specifically explored how to exploit weak labeling to provide feedback at present. Besides, it also analyzed how to provide users with a unified intervention experience.
It is involved in numerous tasks, such as person re-identification, face recognition, 3D point cloud object detection, and object detection.
While considerable current pedestrian re-identification (Re-ID) methods can achieve superior results under the training of a large amount of labeled data, these models cannot produce an exceptional performance as in the experiment when deployed in a natural environment. Moreover, so much data is new in a natural environment because these data have not been in the training set. The trickier part is that new data will constantly accumulate over time, which can cause the model to fail to work. To tackle this problem, Liu~\etal~\cite{liu2019deep} proposed a human-in-cycle model based on reinforcement learning, which released the limitation of pre-labeling and upgraded the model through continuously collected data. The goal is to minimize human annotation work meanwhile maximizing the performance of Re-ID.
In addition to directly using reinforcement learning for dynamic learning, researchers also pay attention to expanding and refining data on a new task.
Facial expression recognition is an exciting task in CV, which is of great help to sentiment analysis and behavior analysis tasks. Traditional facial expression recognition can only deal with the seven simplest facial expressions (\ie happiness, sadness, fear, anger, disgust, surprise, and contempt).
In real life, it is additionally important to deal with more micro-expressions. More specifically, it is an interesting task that builds more refined micro expression processing datasets based on existing expression identification. Butler~\etal~\cite{butler2020human} exploited a micro-expression recognition method based on the \REO{human-in-the-loop} system. This method provides a flexible interface for manual proofreading of automatically processed tags, thereby ensuring the accuracy and usability of the extended dataset.
In addition to directly constructing new datasets, it is also of great significance to explore existing datasets, especially for tasks that are difficult to label, such as target detection tasks, the labeling workload is enormous.
To reduce the labor and time cost of annotation of the bounding box of video objects, Le~\etal~\cite{le2020toward} applied an efficient and straightforward interactive self-annotation framework based on cyclic self-supervised learning. The entire framework consists of automatic model learning and interactive processes.
\p{The automatic learning process makes the model learn faster and speeds up the interaction process.} In the interactive recursive annotation, the detector receives feedback from the human annotator to process the human loop annotation scene. Moreover, to save labeling time, they proposed a new level correction module, which strengthens the utility of neighbor frames by CNN by reducing the distance of annotated frames at each time step.
Based on the framework of Le~\etal, Adhikari~\etal~\cite{adhikari2021iterative} modified the framework to be completed in one stage, and the most significant work of humans in it has become to correct errors instead of performing full annotations, which further improved the user experience.

Using the above two approaches is ineffective for more intricate image tasks, such as 3D point cloud labeling tasks. Because of the limited effect of employing only one stage for labeling, Meng~\etal~\cite{meng2021towards} designed a multi-stage \REO{human-in-the-loop} labeling method based on predecessors.
However, the previous work only started from the perspective of data annotation. \p{It ignored integrating human experience and knowledge into the model to the greatest extent to incorporate human knowledge and intelligence effectively.}
Zhang~\etal~\cite{zhang2021generating} considered specific talents and skills of humans in painting, and these skills cannot be fully quantified as rules and knowledge.
If the model can learn painting skills, it would undoubtedly help \REO{human-in-the-loop} application takes a significant step forward.
They present a data-driven framework for generating comics from digital illustrations.
To further create high-quality comics, these three components are humanely annotated by the artist.
\begin{figure}
    \centering
    \includegraphics[width=0.88\linewidth]{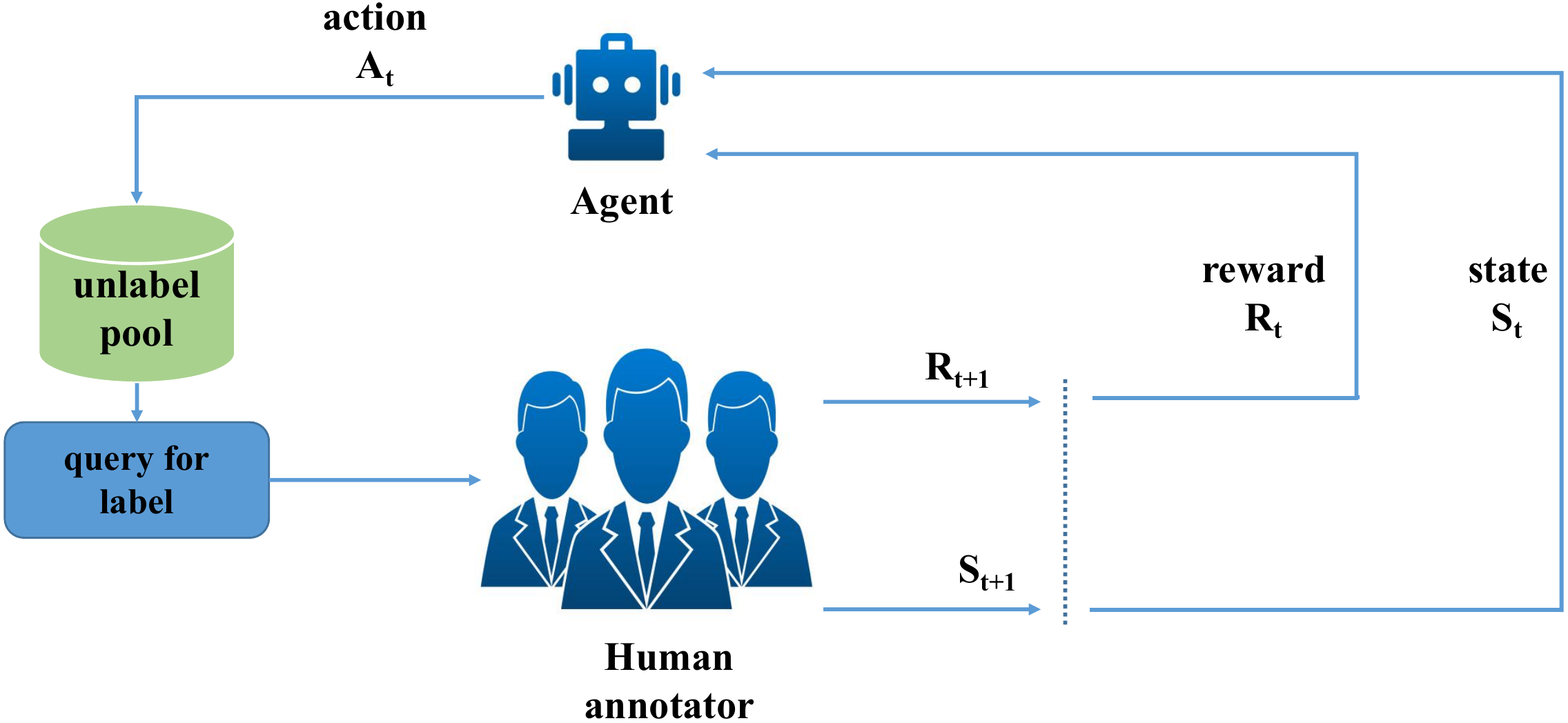}
    \caption{The \REO{human-in-the-loop} framework based on reinforcement learning.
    }
    \label{Dp3}
\end{figure}

\subsection{Iterative Labeling}
At present, there is still a high degree of coupling between deep learning tasks and data processing, and the performance of deep learning largely depends on the quality of the data.
A large amount of high-quality labeled data is needed for a new task to obtain better performance.
However, labeling large-scale data requires a lot of labor and takes a long time, while many iterations of tasks cannot afford such a cost and wait such a long time.
Unlike weak annotate and automatically annotate, \REO{human-in-the-loop}-based methods emphasize finding the essential samples that play a decisive factor in new sample data.

Unlike the data annotation mentioned above in \ref{subsec:Data Annotation}, data iterative labeling pays more additional attention to user experience, not just directly allowing users to perform data annotation.
From annotation to iterative labeling, the goal has been changed in the following two aspects: to focus on adding knowledge and experience to the learning system. The other is to focus on the interaction with users.

Yu~\etal~\cite{yu2015lsun} utilized a partially automated labeling scheme for annotation, which free up human labor by using deep learning of \REO{human-in-the-loop}.
This constitutes the basic prototype of simple iterative annotation.
Recently, with the proliferation of reinforcement learning, Liu~\etal~\cite{liu2019deep} developed a representative \REO{human-in-the-loop} system based on reinforcement learning, which applied reinforcement learning to carry out iterative labeling. A typical framework of \REO{human-in-the-loop} for reinforcement learning is shown in Fig.~\ref{Dp3}. This novel attempt extends the practical usability of \REO{human-in-the-loop} to the field of reinforcement learning for the first time, which brings a valuable contribution to the \REO{human-in-the-loop} community.
In addition to implementing the simple manual intervention, they take person Re-Identification as a research task and explore how to minimize human annotation work while optimizing the performance of Re-ID. Fan~\etal~\cite{fan2019interactive} set out to solve the data challenge in the scheme of network anomaly detection to allow users to intervene in data labeling rather than implement simple user labeling.
They introduce a new intelligent labeling method, and the method combines active learning and visual interaction to detect network abnormalities through the iterative labeling process of users.
The difference is that they began to pay attention to the connection between the algorithm and the visual interface, and the algorithm and the optical interface are tightly integrated.

\section{Model Training and Inference}
\label{sec:MTI}
In many fields of Artificial Intelligence, such as Natural Language Process (NLP) and Computer Vision (CV), there are a variety of approaches that leverage human intelligence to train and infer experimental results.
For both NLP and CV, related research spans deep learning~\cite{lecun2015deep} techniques and human-machine hybrid methods.
These heuristic methods have taken the diverse quality of human creativity into account to achieve high-quality results.

\begin{figure}[t]
    \centering
    \includegraphics[width=0.88\linewidth]{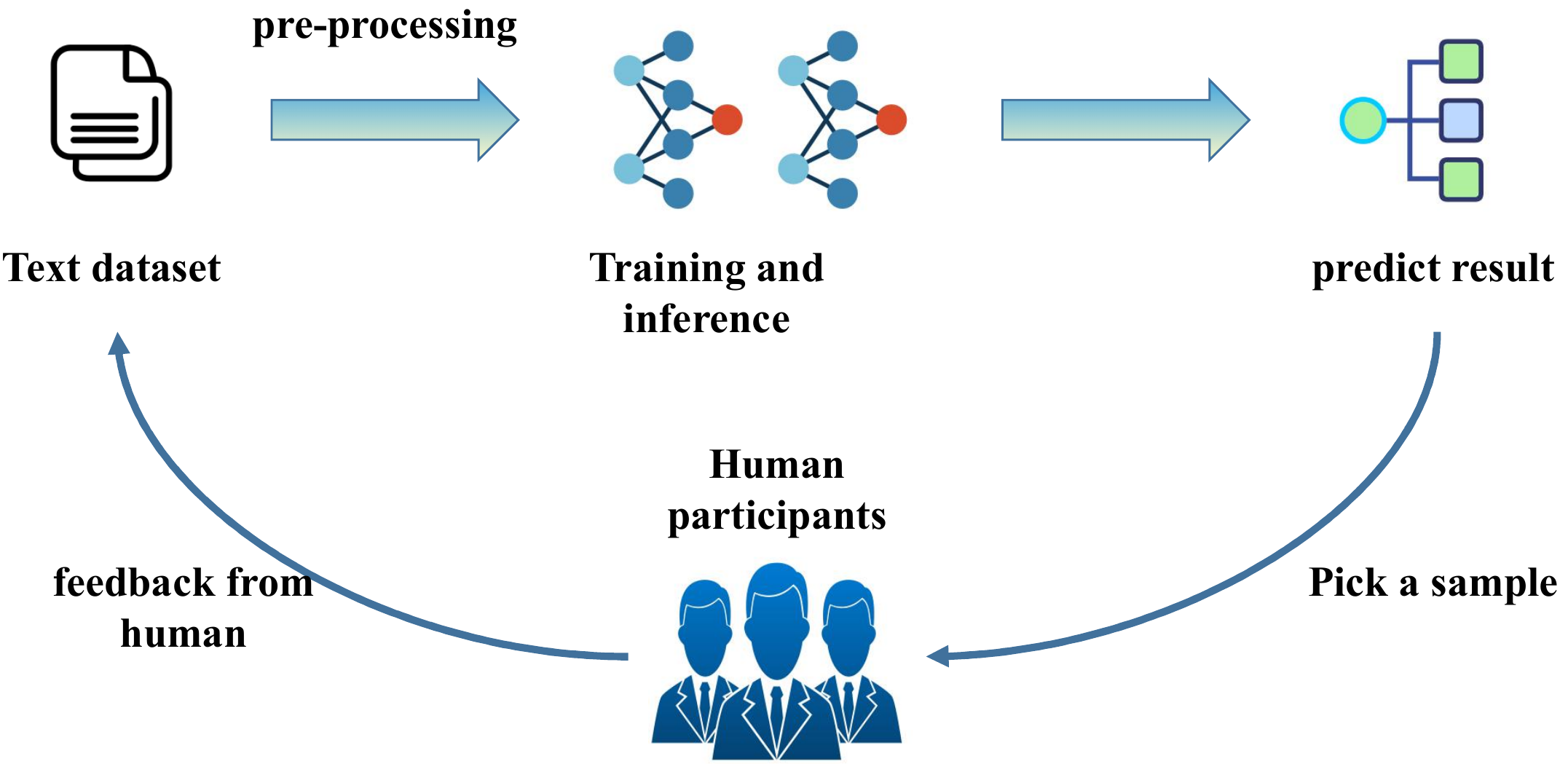}
    \caption{The model training and inferencing workflow of \REO{human-in-the-loop} in Natural Language Processing. The human participants provide various feedback in the stage of model training and inferencing according to specific tasks to boost the performance of NLP models.
    }
    \label{fig_xlw_1}
\end{figure}
\subsection{Natural Language Process}
Fig.~\ref{fig_xlw_1} briefly illustrates the cooperation between the individuals and the model training and inferencing process in the Natural Language Processing Loop. The continuous executive loop develops a more reliable human-AI partnership to a certain extent, contributing to higher accuracy and stronger robustness of the NLP system.

\subsubsection{Text Classification}

Text Classification (TC) is a fundamental NLP task that categorizes a sentence/text into its corresponding category. Karmakharm~\etal \cite{karmakharm2019journalist} propose a rumor classification system\ZJH{. The} core idea of this system is to obtain additional manual feedback from the journalists to retrain a more accurate machine learning model. This framework first exploits a Rumour Classification System to classify collected social media posts and sends \ZJH{these} information back to the journalists. As most state-of-the-art text classification approaches are dominated by the deep neural network~\cite{song2019targeted, bai2020investigating}, which is generally considered as ``black boxes" by end-users, another motivation for the researcher to construct the \REO{human-in-the-loop} framework for TC is to overcome the opaqueness of those models, \ZJH{making} them more explainable. To achieve this purpose, Arous~\etal~\cite{arous2021marta} put forward a hybrid human-AI framework that gives a moral idea to reinforce human reliability in merging human rationales into a deep learning algorithm. Their work presents MARTA, a Bayesian framework that jointly learns and updates the model parameters and human reliability via an iterative way, enabling the learning processes of parameters and human reliability \ZJH{to} benefit from each other until the label and rationales reach agreements.

\subsubsection{Syntactic and Semantic Parsing}

Besides text classification, \REO{human-in-the-loop} approaches for syntactic and semantic parsing are also promising. Syntactic parsing is a process to obtain the valid syntactic structure of input sentences. The goal of \ZJH{semantic parsing} is to map natural language to formal domain-specific semantic representations. A \REO{human-in-the-loop} parsing method~\cite{he2016human} is proposed to improve the parsing accuracy of CCG parsing by employing non-expert to answer simple what-questions generated from the parser's output. These answers are treated as soft constraints when re-training the model. This work is the first attempt at introducing \REO{human-in-the-loop} for syntactic parsing. However, most parsing technologies still face several challenges: (1) the purpose or expression of users can be ambiguous or vague under some circumstances, posing obstacles for them to get the ground truth in one shot, (2) in the real-world scenario, the performance of state-of-the-art parsers are generally not high enough, and (3) since the mainstream neural network-based models are known as ``black-box" \ZJH{that} indicates the lack of explainability, it is difficult for end-users to verify the parsing results independently.
Currently, Yao~ \etal~\cite{yao2019interactive} \ZJH{propose} allowing the semantic parsers system to ask end-users clarification questions and produce an If-Then program simultaneously. Although recent works successfully verified the effectiveness of interactive semantic parsing in practice, they are generally restricted to a specific type of formal language.
\ZJH{Furthermore, }Yao~ \etal~\cite{yao2019model} develop a model-based interactive semantic parsing (MISP) as the general principle for interactive semantic parsing.

\begin{table*}
\centering
\caption{A brief overview of representative works in \REO{human-in-the-loop} NLP. Each row represents one work. Works are sorted by task types (TC: Text Classification. SSP: Syntactic and Semantic Parsing. TS: Text Summarization. QA: Question Answering. SA: Sentiment Analysis). Each column corresponds to a dimension from the two subsections (task, motivation).}
\label{tab_xlw_1}
\scalebox{0.78}{\begin{tabular}{l|ccccc|p{18px}<{\centering}p{18px}<{\centering}p{18px}<{\centering}|c}
\toprule
\multicolumn{1}{c|}{}                                & \multicolumn{5}{c|}{{\color[HTML]{000000} \textbf{Task}}}                                                    & \multicolumn{3}{c|}{{\color[HTML]{000000} \textbf{Motivation}}}                                         \\ \cmidrule(l){2-9}
\multicolumn{1}{c|}{\multirow{-2}{*}{\textbf{Work}}} & \multicolumn{1}{l}{TC} & \multicolumn{1}{l}{SSP}  & TS & \multicolumn{1}{l}{QA} & \multicolumn{1}{l|}{SA} & \multicolumn{1}{l}{Performance} & \multicolumn{1}{l}{Interpretability} & \multicolumn{1}{l}{Usability} & \multicolumn{1}{|c}{\multirow{-2}{*}{\textbf{\REO{Quantitative results}}}} \\ \midrule
\textit{Arous et al. (2021)} \cite{arous2021marta}     & \checkmark                                               &    &    &                        &                         & \checkmark                               & \checkmark                                    &                         &\REO{$0.840->0.960(Acc)$}       \\
\textit{Karmakharm et al. (2019)} \cite{karmakharm2019journalist}& \checkmark                                              &    &    &                        &                         & \checkmark                               &                                      & \checkmark                          &\REO{$--$}      \\
\textit{Yao et al. (2019)} \cite{yao2019interactive}       &                        & \checkmark                           &    &                        &                         & \checkmark                               & \checkmark                                    & \checkmark                        &\REO{$0.640->0.968(Acc)$}        \\
\textit{Yao ZiYu et al. (2019)} \cite{yao2019model}  &                        & \checkmark                           &    &                        &                         & \checkmark                               & \checkmark                                    &                           &\REO{$0.615->0.729(Acc)$}       \\

\textit{Ziegler et al. (2019)} \cite{ziegler2019fine}                        &                         &    & \checkmark  &                        &                         & \checkmark                               &                                      &                             &\REO{$--$}     \\
\textit{Stiennon et al. (2020)} \cite{stiennon2020learning}                        &                         &    & \checkmark  &                        &                         & \checkmark                               &                                      &                           &\REO{$--$}       \\
\textit{Hancock et al. (2019)} \cite{hancock2019learning}                          &                         &    &    & \checkmark                      &                         & \checkmark                               &                                      &                          &\REO{$0.447->0.463(Acc)$}        \\
\textit{Wallace et al. (2019)} \cite{wallace2019trick}                          &                         &    &    & \checkmark                      &                         & \checkmark                               &                                      & \checkmark                           &\REO{$0.4->0.6(Acc)$}     \\
\textit{Liu et al. (2021)} \cite{liu2021and}                              &                         &    &    &                        & \checkmark                       &                                 & \checkmark                                    & \textbf{}                &\REO{$0.71->0.82(Precision)$}        \\ \bottomrule
\end{tabular}}
\end{table*}

\subsubsection{Text Summarization}

Besides applying a \REO{human-in-the-loop} framework to topic modeling, researchers also use \ZJH{it} to generate new texts. Text Summarization (TS) generates a shorter version of a given sentence/text while preserving its meaning~\cite{chopra2016abstractive}.
In recent years, there have been some significant breakthroughs in this field. For instance, Ziegler~\etal~\cite{ziegler2019fine} fine-tune pre-trained language models with reinforcement learning by exploiting a reward model trained from human preferences. Then the model is used to generate summaries over Reddit TL, DR, and CNN/DM datasets. However, one limitation of their framework is that there are low agreement rates between labelers and researchers.
\ZJH{Stiennon~\etal~\cite{stiennon2020learning} propose to gather a dataset composed of human preferences between pairs of summaries as the first step}. Then the prediction of the human-preferred summary is generated by a reward model (RM) trained via supervised learning. Lastly, the score produced by the RM is maximized as much as possible by a policy trained via reinforcement learning (RL). Their method ensures a relatively higher labeler-researcher agreement through the above steps and successfully separates the policy and value networks.

\subsubsection{Question Answering}

Recently, various \REO{human-in-the-loop} related frameworks have been designed to apply dialogue and Question Answering (QA). The purpose of this task is to allow chatbots/agents to have a conversation with users. These \REO{human-in-the-loop} dialogue intelligent systems can be bifurcated into two main categories: online feedback loop and offline feedback loop~\cite{wang2021putting}. For the online feedback loop, human feedback is utilized to update the model continuously. Compared with traditional approaches that mismatch the training set and online use case for dialogue systems, researchers have demonstrated that the application of online reinforcement learning can improve the model with human feedback. For instance, a lifetime learning framework is proposed by Hancock~\etal~\cite{hancock2019learning}. The self-feeding mechanism in this framework enables the chatbot to generate new examples when the conversation with users goes well, and these new examples are exploited to re-train itself continuously. For the offline feedback loop, a large set of human feedback needs to be collected as a training set, then this training set is used to update the model. \ZJH{For instance, Wallace~\etal~\cite{wallace2019trick} employ ``trivia enthusiasts" to creatively generate adversarial examples that can confuse their QA system.} These examples are finally implemented for negative training. Since some of the end-user feedback can be misleading, offline methods \ZJH{is} more appropriate for improving the robustness of the model.

\subsubsection{Sentiment Analysis}

Sentiment Analysis (SA) is one of the attractive research branches of Opinion Mining (OM). \ZJH{The research scope of SA is the computational study of individuals' opinions and attitudes toward entities mentioned in a text}. The entities generally refer to individuals or events. Recently numerous neural network-based approaches have been widely utilized and demonstrated their effectiveness in solving sentiment analysis tasks~\cite{song2019targeted, xiao2020targeted, xiao2020multi}. Most deep learning-based methods for SA use accuracy and F1-score as evaluation metrics. Since these metrics can only evaluate the predictive performance, they lack the mechanisms to explain when and why the sentiment models give false predictions in run-time~\cite{nushi2018towards}. Liu~\etal~\cite{liu2021and} \ZJH{introduce} an explainable \REO{human-in-the-loop} SA framework for sentiment analysis \ZJH{task}. The execution of their framework \ZJH{is} segmented into three steps: First, the \REO{human-in-the-loop} SA model analyzes local feature contributions. This goal is achieved by executing a data perturbation process. Next, local features are aggregated to calculate the explainable global-level features and humans participate in this loop to assess the relevance of the top-ranked global features to the ground truth and report the errors they find in this process. Finally, the system calculates an erroneous score based on global-level and local-level sentimental features for each instance. Scores higher than a specific threshold are indicated as wrong predictions.

\REO{\subsubsection{Summarization for Human-in-the-Loop in NLP}}

A brief overview of representative works in \REO{human-in-the-loop} NLP is shown in Table~\ref{tab_xlw_1}.
\ZJH{For most of the surveyed papers above, their original purpose is to apply \REO{human-in-the-loop} techniques to various NLP tasks for better performance. }
The effectiveness of the approaches proposed by these surveyed papers is evaluated via multiple metrics. The experimental results in the documents we \ZJH{investigate} show that a relatively small set of human feedback can dramatically boost model performance. For instance, the \REO{human-in-the-loop} technique improves the classification accuracy for text classification~\cite{karmakharm2019journalist}. Similar situations occur in dialogue and question answering where the QA systems have higher ranking metric hits \cite{hancock2019learning}. Besides, \REO{human-in-the-loop} techniques have also enhanced the model's robustness and generalization \cite{stiennon2020learning}. In addition to improving model performance, some studies have demonstrated that \REO{human-in-the-loop} methods enable models to be more interpretable and usable in solving NLP problems. For instance, Arous~\etal~\cite{arous2021marta} incorporate human rationales into an attention-based Bayesian framework reasonably while weighing worker reliability, thus providing a more human-understandable interpretation of classification results and enhancing the model performance at the same time. Liu~\etal~\cite{liu2021and} chose uni-grams as the explainable feature for LIME~\cite{ribeiro2016should}; thus the proposed system allows the end-users to understand better the overall contribution of each word to the final sentiment classification made by the model. Wallace~\etal~\cite{wallace2019trick} invite ``trivia enthusiasts" to creatively generate specific adversarial questions that confuse the intelligent question answering system. These questions can be treated as probes further to explore the inherent characteristics of the underlying model behaviors.

\subsection{Computer Vision}

\begin{figure}[t]
    \centering
    \includegraphics[width=0.88\linewidth]{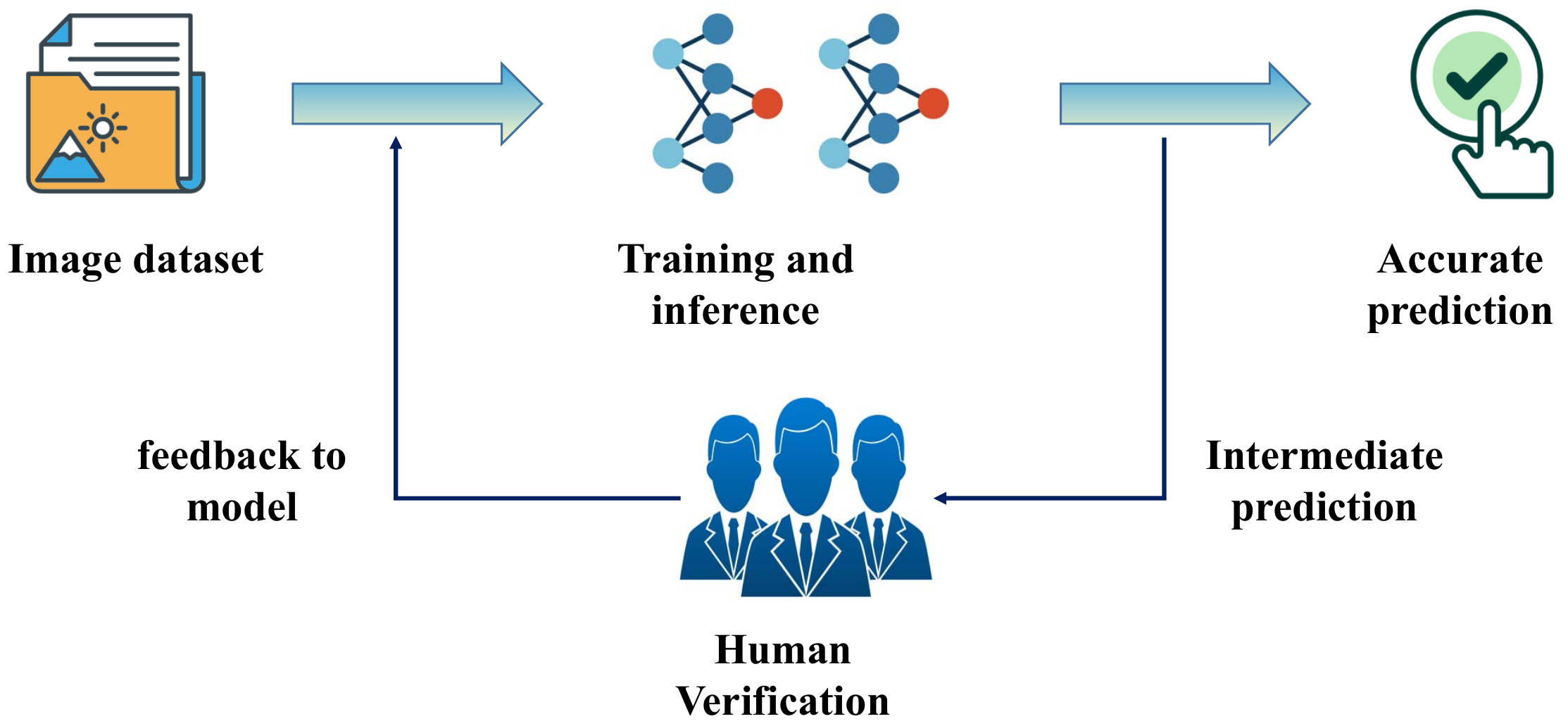}
    \caption{Overview of general \REO{human-in-the-loop} frameworks for model training and inferencing in Computer Vision.
    }
    \label{fig_xlw_2}
\end{figure}

In recent years, neural network-based Deep Learning methods (DL) have emerged as the state-of-the-art technique for performing many computer vision tasks~\cite{WU2021436, wu2020fast}. To further improve the performance of these methods, feedback from humans has been integrated into the deep learning architecture to make the whole system more intelligent in solving complex cases that can not be handled politely by the model. A typical \REO{human-in-the-loop} framework for Computer Vision is outlined in Fig.~\ref{fig_xlw_2}.

\subsubsection{Object Detection}

Object detection, as one of the most fundamental and challenging problems in computer vision~\cite{girshick2015fast}, has received significant attention in recent years~\cite{zou2019object}. The goal of object detection is to detect instances of visual objects of a specific class (such as individuals, vehicles, or other creatures) in digital images.
Yao~\etal~\cite{yao2012interactive} point out that iterations between queries may be expensive \ZJH{and} time-consuming, making it unrealistic for executing interaction with end-users. They present an interactive object detection architecture to employ individuals to correct a few annotations proposed by a detector for the un-annotated image with the maximum predicted annotation cost. However, it is still difficult to detect some occluded objects, tiny objects, and blurred objects \ZJH{by} these approaches. Madono~\etal~\cite{madono2020efficient} put forward an efficient \REO{human-in-the-loop} object detection framework composed of bi-directional deep SORT~\cite{wojke2017simple} and annotation-free segment identification (AFSID). Humans' role in this architecture is to verify the object candidates that bi-directional deep SORT can not detect automatically. Then train the model over the supplementary objects annotated by individuals.

\subsubsection{Image Restoration}
Image restoration (IR) aims to recover the preliminary version of damaged images~\cite{andrews1977digital}. The image inpainting frameworks proposed by previous studies can be bifurcated as exemplar-based~\cite{criminisi2003object} approaches and deep learning-based~\cite{liu2018image} methods. Although profound learning-based works are the mainstream and show decent results, neural network-based approaches constantly suffer from over-fitting when only a relatively small training set is available on a large dataset. Besides, \ZJH{in real-world application,} the restored images are often filled with unknown artifacts like uneven texture or monotone color due to the missing crucial semantic information in severely corrupted areas. Weber~\etal~\cite{weber2020draw} propose an interactive machine learning system for image restoration based on Deep Image Prior (DIP)~\cite{ulyanov2018deep}.
\ZJH{Their proposed \REO{human-in-the-loop} framework embeds human knowledge into the training process by the following steps. }
Initially, the images from the dataset are sent to the automated DIP for preliminary restoration.
\ZJH{Secondly, the operators actively refine the images via a pre-designed user interface.
Thirdly, the refined images are sent back to the input of DIP again for further polishing. Finally, the whole loop continues until the restoration reaches the user's expectation.} In the field of Electron Microscopy, one drawback of automation is that it generally ignores the expertise of the microscopy user that comes with manual analysis. To alleviate such a challenging problem, Roels~\etal~\cite{roels2019human} propose a hybrid \REO{human-in-the-loop} system that incorporates expert microscopy knowledge with the power of large-scale parallel computing to enhance the Electron Microscopy image quality by exploiting image restoration algorithms.

\subsubsection{Image Segmentation}

\ZJH{Image segmentation (i.e. semantic segmentation) is a crucial step in most image studies. Image segmentation (IS) aims at assigning a class label to each pixel in the image~\cite{minaee2021image}.} This field has recently become explosive popularity because it plays a crucial role in a wide range of computer vision applications~\cite{badrinarayanan2017segnet}. However, few works explore how to effectively expose failures of ��top-performing �� semantic segmentation models and rectify the models by utilizing such counter-examples reasonably. Wang~\etal~\cite{wang2020efficiently} \ZJH{present} a two-step hybrid system with human efforts for troubleshooting pixel-level image labeling models. The hybrid system first automatically picks up un-labeled images from a large pool. These selected unlabelled images are used to compose an unlabeled set, which is the most informative in exposing the weaknesses of the target model. To reduce the number of false positives, individuals filter the unlabeled set to obtain a smaller straight set. In the second step, they fine-tune and re-train the target model to study from the counter-examples contained in the refining set without ignoring previously seen examples. Data annotation is always complicated and expensive in the medical image process domain~\cite{taleb2021multimodal}. Ravanbakhsh~\etal~\cite{ravanbakhsh2020human} introduce a training protocol based on combining the conditional Generative Adversarial Network (cGAN) and human workers interactively. For complex cases, human experts are responsible for annotating them. These newly annotated images are used to continue the training and inference procedure.

\begin{table*}[t]
\centering
\caption{A brief overview of representative works in \REO{human-in-the-loop} CV. Each row represents one work. Works are sorted by their task types (OD: Object Detection. IR: Image Restoration. IS: Image Segmentation. IE: Image Enhancement. VOS: Video Object Segmentation). Each column corresponds to a dimension from the two subsections (task, motivation).}\label{tab_xlw_2}
\scalebox{0.78}{\begin{tabular}{l|ccccc|ccc|c}
\toprule
\multicolumn{1}{c|}{}                                & \multicolumn{5}{c|}{{\color[HTML]{000000} \textbf{Task}}}                           & \multicolumn{3}{c|}{{\color[HTML]{000000} \textbf{Motivation}}}                                         \\ \cmidrule(l){2-9}
\multicolumn{1}{c|}{\multirow{-2}{*}{\textbf{Work}}} & \multicolumn{1}{l}{OD} & \multicolumn{1}{l}{IR} & IS & IE & \multicolumn{1}{l|}{VOS} & \multicolumn{1}{l}{Performance} & \multicolumn{1}{l}{Interpretability} & \multicolumn{1}{l}{Usability} &\multicolumn{1}{|c}{\multirow{-2}{*}{\textbf{\REO{Quantitative results}}}}\\ \midrule
\textit{Yao et al. (2012)}~\cite{yao2012interactive}        & \checkmark                      &                        &    &    &                          & \checkmark                               &                                      &                             &\REO{$--$}   \\
\textit{Madono et al. (2020)}~\cite{madono2020efficient}      & \checkmark                      &                        &    &    &                          & \checkmark                               &                                      &                             &\REO{$--$}   \\
\textit{Roels et al. (2019)}~\cite{roels2019human}       &                        & \checkmark                      &    &    &                          & \checkmark                               &                             \checkmark         &                             &\REO{$--$}  \\
\textit{Weber et al. (2020)}~\cite{weber2020draw}       &                        & \checkmark                      &    &    &                          & \checkmark                               &                                      &                            &\REO{$0.2816->0.2227(DSSIM)$}    \\
\textit{Wang et al. (2020)}~\cite{wang2020efficiently}        &                        &                        & \checkmark  &    &                          & \checkmark                               &                                      &                           &\REO{$0.1365->0.4233(mIoU)$}    \\
\textit{Ravanbakhsh et al. (2020)}~\cite{ravanbakhsh2020human} &                        &                        & \checkmark  &    &                          & \checkmark                               &                                      &                             &\REO{$0.645->0.846(Acc)$}  \\
\textit{Murata et al. (2019)}~\cite{murata2019automatic}      &                        &                        &    & \checkmark  &                          & \checkmark                               &                                      &                              &\REO{$--$}   \\
\textit{Fischer et al. (2020)}~\cite{fischer2020nicer}     &                        &                        &    & \checkmark  &                          & \checkmark                               &                                      &                            &\REO{$--$}    \\
\textit{Benard et al. (2017)}~\cite{benard2017interactive}      &                        &                        &    &    & \checkmark                        & \checkmark                               &                                      & \checkmark                           &\REO{$0.504->0.822(IoU)$}   \\
\textit{Oh et al. (2019)}~\cite{oh2019fast}          &                        &                        &    &    & \checkmark                        & \checkmark                               &                                      & \checkmark
\textbf{}                     &\REO{$0.555->0.691(AUC)$}  \\ \bottomrule
\end{tabular}}
\end{table*}

\subsubsection{Image Enhancement}
As one of the challenging issues in computer vision, the purpose of Image Enhancement (IE) is to process an image and generate a \ZJH{new and improved one that is more suitable} for a specific application~\cite{shukla2017review}. The research field of image enhancement has attracted ample attention from researchers in recent years, especially after the emergence of deep neural network algorithms~\cite{fu2018image}. However, most current frameworks have ignored the user preferences and experiences, enhancing the image only via a black-box style, which can leave end-users with sub-optimal results that are not suitable for their specific taste. Murata~\etal~\cite{murata2019automatic} take user preference into account. \ZJH{The user first provides an example to his system;} the image enhancement functions in the framework are applied to the example image via randomly selected parameters. Several objective photos are produced, and the end-user needs to score each of the images. Then the RankNet~\cite{burges2005learning} is exploited to learn the user's preference from these scores.
During the learning process, based on the scores given by users, parameters are optimized to make the generated enhanced images suitable for the taste of users. Fischer~\etal~\cite{fischer2020nicer} devise Neural Image Correction and Enhancement Routine (NICER). \ZJH{A component called Image Manipulator in the NICER first exploits a series of learned image operations (e.g. contrast, brightness)} with variable magnitude onto the original image provided by users. Another module named Quality Assessor is followed to evaluate the final enhancement quality by generating related scores. This system iteratively optimizes the parameters of the image enhancement functions to maximize the scores given by the Quality Assessor. Users can modify the parameters of the Image Manipulator before, during, and after the optimization process, guiding the optimization procedure towards more satisfying local optima.

\subsubsection{Video Object Segmentation}
The goal of \ZJH{Video Object Segmentation} (VOS) is to segment a particular object instance in the entire video sequence of the object mask on a manual or automatic first frame~\cite{yao2020video}. This research area has become popular in the computer vision community~\cite{caelles2017one}. Since videos have intrinsic characteristics such as motion blur, bad composition, occlusion, etc., it is harder for fully automatic approaches to segment more complex sequences accurately.
Employing user \ZJH{to} input for each frame is unrealistic due to its expensive costs and time consumption. Thus, the \REO{human-in-the-loop} framework is adopted for solving such problems. Benard~\etal~\cite{benard2017interactive} introduce a novel \ZJH{interactive video object segmentation} method based on~\cite{xu2016deep}. The core idea of their \REO{human-in-the-loop} framework is to utilize the current segmentation mask as an additional input. A practical framework for interactive segmentation scenario is designed by Oh~\etal~\cite{oh2019fast}, named Interaction-and-Propagation Networks (IPN). The IPN is composed of two modules and the critical architecture of these two modules is deep convolutional neural networks. The primary operations of these two modules are interaction and propagation, respectively. Individuals are allowed to interact with the proposed model several times; meanwhile, the feedback is provided in scribbles on multiple frames during this interactive procedure.

\REO{\subsubsection{Summarization for Human-in-the-Loop in CV}}
A brief overview of representative studies in \REO{human-in-the-loop} CV is displayed in Table~\ref{tab_xlw_2}. It can be observed from Table~\ref{tab_xlw_2} that the motivation of all the surveyed \REO{human-in-the-loop} works for computer vision is to boost the model performance. From the experiment results of all these surveyed papers, although the evaluation criteria are different, the system that incorporates the \REO{human-in-the-loop} method performs better than without combining it. Taking Madono~\etal~\cite{madono2020efficient} as an example, they conduct experiments for pedestrian detection and the results have proven at least two advantages for this task: For one thing, the proposed approach boosts the recall rate by \ZJH{11$\%$} at most over deep SORT.
For \ZJH{the other}, the amount of unlabeled samples that need manual annotation is decreased by  \ZJH{67$\%$} at most compared with bi-directional deep SORT without AFSID, which dramatically improves the overall model performance. Associating the contents in Table~\ref{tab_xlw_1}, this phenomenon in CV is similar to NLP, which demonstrates that the core motivation of almost all \REO{human-in-the-loop} studies in both CV and NLP serves the purpose of boosting model performance. We also notice that in Table~\ref{tab_xlw_2}, only one work~\cite{roels2019human} tries to bring interpretability for the model. Roels~\etal~\cite{roels2019human} have validated the potential enhancements \ZJH{that} DenoisEM can provide in 3D EM image interpretation by denoising SBF-SEM image data of an Arabidopsis thaliana root tip. Besides, \REO{human-in-the-loop} conception can also improve the usability of CV models. For instance, Madono~\etal~\cite{madono2020efficient} have proven that their framework is advantageous/practical in scenarios where obtaining annotations is a costly affair. Oh~\etal~\cite{oh2019fast} validate the usefulness and robustness of their Interaction-and-Propagation Networks with real interactive cutout use-cases.
\REO{Hudec~\etal~\cite{hudec2021classification} propose a new classification of aggregate functions in terms of mixed behavior by the variability of ordinal sums of associative and disjunctive functions, which play an important role for model fusing the knowledge of domain experts.}

\section{System construction and Application}
\label{sec:SCA}
\begin{table*}[t]
\centering
\caption{An overview of representative applications for software-based \REO{human-in-the-loop} systems. (SECS: Security Systems. CP: Code Production. SIMS: Simulation System. SE: Search Engine.)}
\label{tab_syx_1}
\scalebox{0.60}{\begin{tabular}{l|ccccc|c|cccc|l|c}
\toprule
\multicolumn{1}{c|}{\multirow{2}{*}{Systems}}                                                                & \multicolumn{5}{c|}{Application} & \multirow{2}{*}{Year} & \multicolumn{4}{c|}{Role of human}            & \multicolumn{1}{c}{\multirow{2}{*}{Description}}      & \multicolumn{1}{|c}{\multirow{2}{*}{\REO{Quantitative results}}}        \\ \cline{2-6} \cline{8-11}
                                                                       \multicolumn{1}{c|}{}              & SECS  & CP  & SIMS & SE & Others &                       & Supervisor & Supervisee & Collaborator & User &                                      \multicolumn{1}{c}{}      \\ \hline
Brostoff \& Sasse~\cite{brostoff2001safe}                   & \checkmark     &     &      &    &        & 2001                  &            & \checkmark          &              & \checkmark    & Security system for traditional tasks   &\REO{$--$}    \\
Cranor~\cite{cranor2008framework}                             & \checkmark     &     &      &    &        & 2008                  &            & \checkmark          &              & \checkmark    & Security system for traditional tasks     &\REO{$--$}  \\
MacHiry~\etal~\cite{machiry2013dynodroid}       &       & \checkmark   &      &    &        & 2013                  &            &            & \checkmark            &      & Software testing                       &\REO{$47\%$}    \\
Kovashka~\etal~\cite{kovashka2015whittlesearch} &       &     &      & \checkmark  &        & 2015                  &            &            &              & \checkmark    & Image-based searching engine           &\REO{$--$}    \\
Louis Rosenberg~\cite{rosenberg2016artificial}                 &       &     &      &    & \checkmark      & 2016                  &            &            & \checkmark            &      & Crowdsourcing                &\REO{$--$}              \\
Yan~\etal~\cite{shoshitaishvili2017rise}        &       & \checkmark   &      &    &        & 2017                  & \checkmark          &            & \checkmark            &      & Software testing                 &\REO{$53.45\%$}          \\
Wogalter~\cite{wogalter2018communication}                    & \checkmark     &     &      &    &        & 2018                  &            &            & \checkmark            & \checkmark    & Security system for traditional tasks    &\REO{$--$}   \\
MA~\cite{ma2018towards}                                     &       & \checkmark   &      &    &        & 2018                  & \checkmark          &            &              &      & AI model optimizing in training        &\REO{$60\% (reduction)$}    \\
Salam~\etal~\cite{salam2019human}               &       & \checkmark   &      &    &        & 2019                  & \checkmark          &            &              &      & AI model optimizing in testing           &\REO{$--$}  \\
Plummer~\etal~\cite{plummer2019give}          &       &     &      & \checkmark  &        & 2019                  &            &            &              & \checkmark    & Image-based searching engine            &\REO{$82.28\% (Acc)$}   \\
Fredrik Wrede~\cite{wrede2019smart}      &       &     &      &    & \checkmark      & 2019                  & \checkmark          &            & \checkmark            &      & Stochastic gene regulatory              &\REO{$--$}    \\
Singh and Mahmoud~\cite{singh2020human}                       & \checkmark     &     &      &    &        & 2020                  &            & \checkmark          & \checkmark            & \checkmark    & Security system for traditional tasks   &\REO{$75.43\% (Acc)$}   \\
Demartini\etal~\cite{demartini2020human}        & \checkmark     &     &      &    &        & 2020                  & \checkmark          &            & \checkmark            &      & Security system for modern tasks         &\REO{$--$}  \\
 ODEKERKEN \&  BEX~\cite{odekerken2020towards}              & \checkmark     &     &      &    &        & 2020                  &            &            & \checkmark            &      & Security system for modern tasks        &\REO{$--$}   \\
Bohme~\etal~\cite{bohme2020human}               &       & \checkmark   &      &    &        & 2020                  & \checkmark          &            &              &      & Program repairing                      &\REO{$75\% (Acc)$}    \\
Renner \cite{renner2020designing}                                  &       & \checkmark   &      &    &        & 2020                  & \checkmark          &            & \checkmark            &      & AI model optimizing in the whole steps   &\REO{$--$}  \\
Davidson~\etal~\cite{davidson2021improving}            &       &     & \checkmark    &    &        & 2021                  &            &            & \checkmark            & \checkmark    & Simulation system for decision making    &\REO{$--$} \\
Demirel~\cite{demirel2020digital}                            &       &     & \checkmark    &    &        & 2020                  &            &            &              & \checkmark    & Simulation system for process forecast                  &\REO{$--$}   \\
Metzner~\etal~\cite{metzner2020system}         &       &     & \checkmark    &    &        & 2020                  &            &            &              & \checkmark    & Simulation system for procedure control  &\REO{$--$} \\
Polisetty \& Avinesh~\cite{polisetty2020information}         &       &     &      & \checkmark  &        & 2020                  &            &            &              & \checkmark    & Searching engine                       &\REO{$--$}    \\
Zhu~\etal~\cite{zhu2020easierpath}             &       &     &      &    & \checkmark      & 2020                  & \checkmark          &            & \checkmark            &      & Renal Pathology                      &\REO{$57\% (reduction)$}         \\
 Li~\etal ~\cite{li2020explanations}             &       &     &      &    & \checkmark      & 2020                  &            &            & \checkmark            &      & Model checking                  &\REO{$--$}            \\ \bottomrule
\end{tabular}}
\end{table*}

\REO{Previously, we summarized and reviewed \p{the \REO{human-in-the-loop} workaround of} `data processing' and `model training and inference'. \p{Moreover}, we \p{noticed} that some researchers focus on how to build human-centric applications in \p{a broad spectrum of} application scenarios.
\p{We selected representative studies from four application scenarios of Security Systems, Code Production Tools, Simulation Systems, and Search Engines to summarize, as shown in Table~\ref{tab_syx_1}.}}

\begin{figure}[t]
    \centering
    \includegraphics[width=0.88\linewidth]{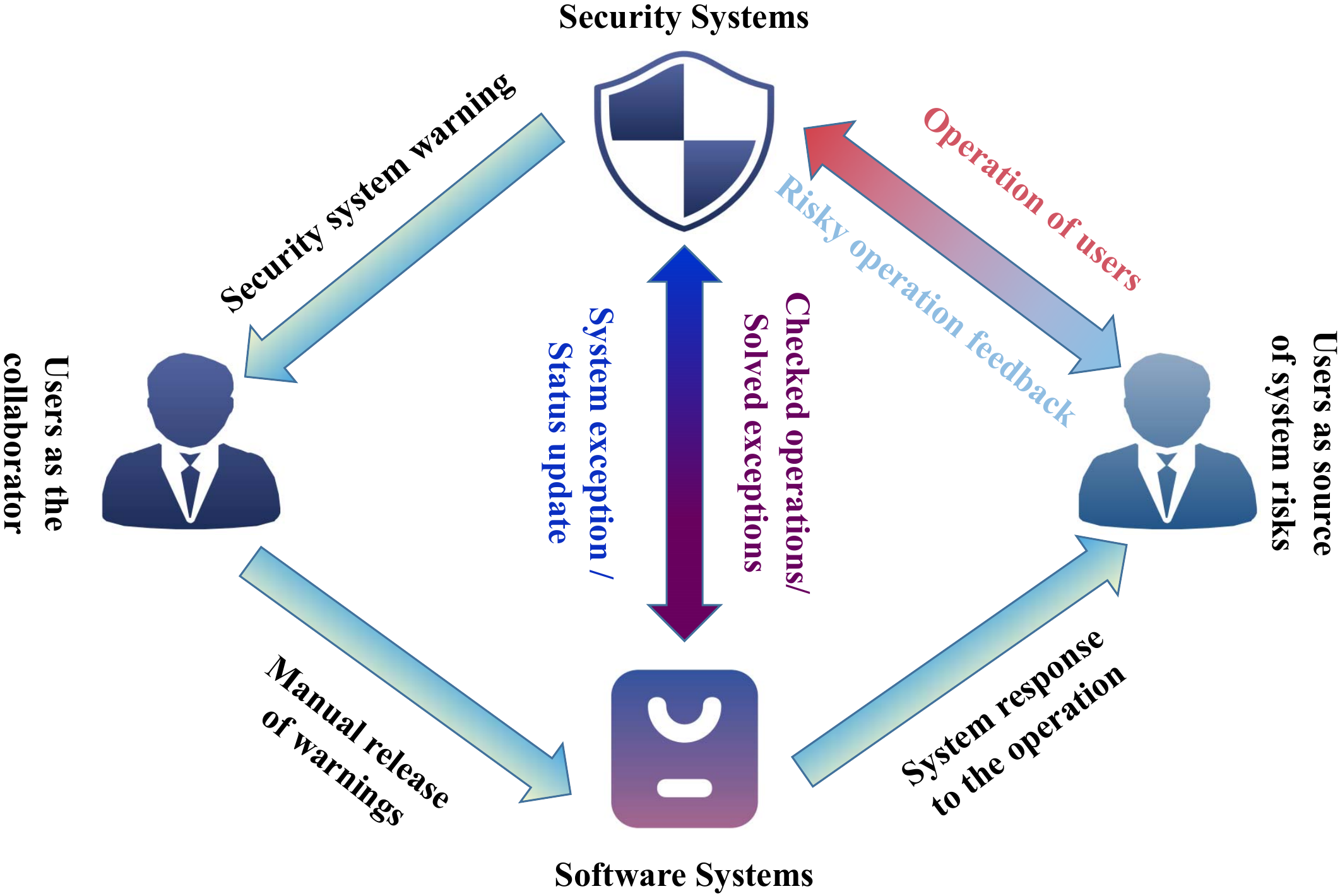}
    \caption{The workflow of the \REO{human-in-the-loop} security system.}
    \label{fig_syx_1}
\end{figure}

\REO{\subsection{Security System}}
\REO{
The security system is an application system with \p{solid} coupling with people (Fig.~\ref{fig_syx_1}). \p{With the development of \p{deep learning algorithm}, researchers begin to explore more effective human-machine cooperation modes to replace the manual intervention process gradually.}
Humans usually fail to execute certain events due to some objective reasons (\p{\ie}inattentiveness, increasing age, nonproficiency).
At present, many safety systems are highly coupled to people, which \p{would} lead to the continuous amplification of errors.
To address this challenge, humans should maximize the chances of operators successfully performing their safety-critical functions.
To this end, Cranor~\cite{cranor2008framework} \p{proposed} a \REO{human-in-the-loop}-based reasoning framework, which \p{provided} a systematic approach to identifying potential causes of human failures, thereby helping system designers use the framework to identify problem areas and proactively address defects before building systems.
Similarly, \p{in order} to stop operator error in time, Singh et al. ~\cite{singh2020human} started by addressing the operational safety of nuclear power plants (NPP) and the commercial aviation industry.
They first used the interaction between operator and control panel (HMI) state, where a visual feedback loop is used to predict the near-term expected state.
\p{The HMI state sequence prediction is transformed into a language translation problem and finely modeled in a \REO{human-in-the-loop} fashion.
The proposed method is validated by a dataset of natural factory scenes, demonstrating the effectiveness of the method.}
}

\begin{figure}[t]
    \centering
    \includegraphics[width=0.88\linewidth]{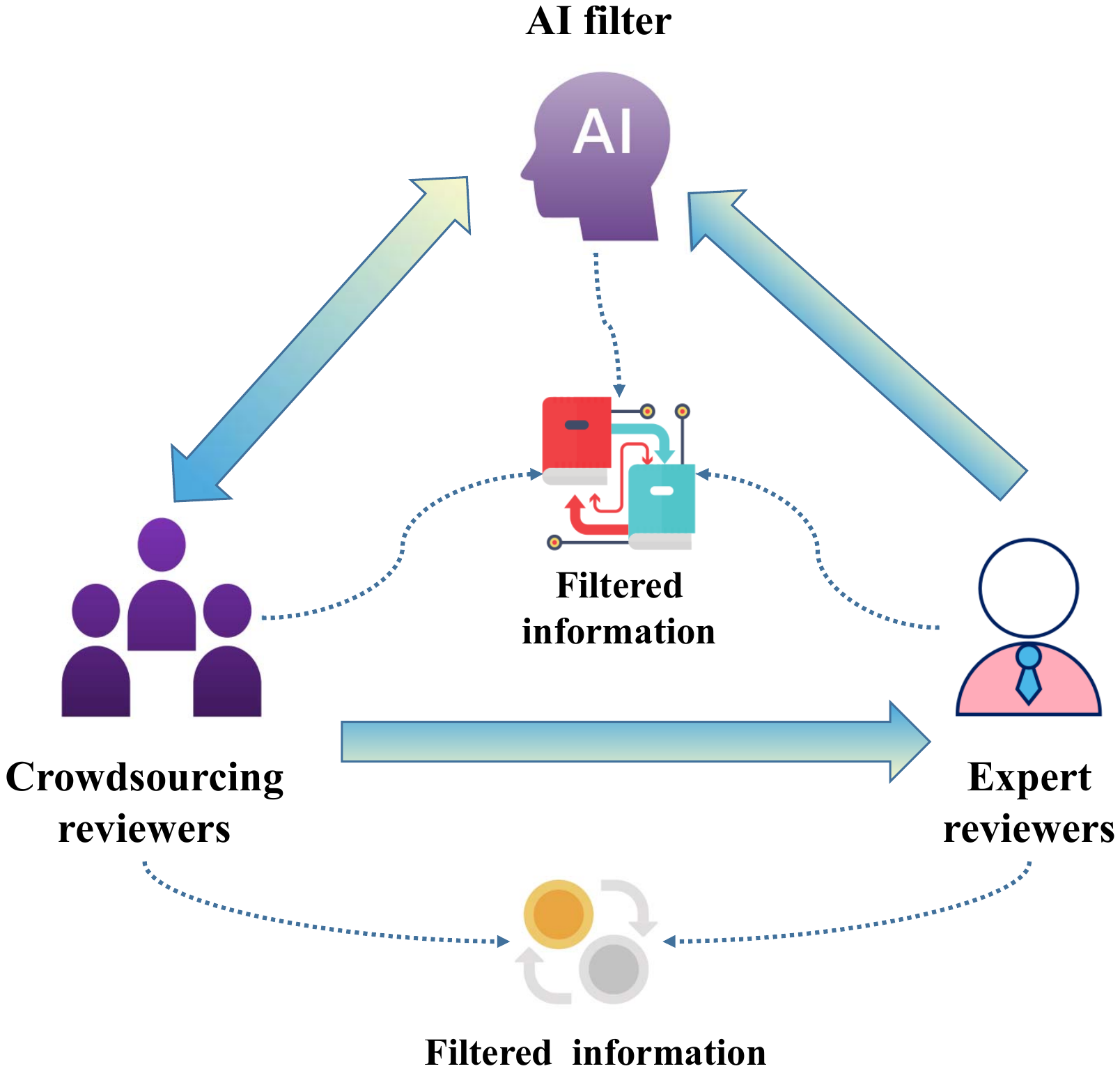}
    \caption{The workflow of the \REO{human-in-the-loop} security system in modern tasks.}
    \label{fig_syx_2}
\end{figure}

The workflow of the \REO{human-in-the-loop} security system can be summarized as Fig.~\ref{fig_syx_2} in which the artificial intelligence algorithm and humans (experts or crowd-sourcing workers) collaborate all together.
\REO{Demartini~\etal~\cite{demartini2020human} \p{devised} a solution to combat online misinformation by combining machine learning algorithms, crowd-sourcing workers and experts, \p{and} their introduction to combining automatic and manual fact-checking methods to the challenges and opportunities in combating the spread of online misinformation.
They \p{pointed out} that by addressing the main question of ``who should do what", the cost and performance of the three roles can be balanced, along with the improvement of credibility and safety of the \REO{human-in-the-loop} system.
For the same purpose, ODEKERKEN~\etal~\cite{odekerken2020towards} \p{introduced} a proxy framework for transparent \REO{human-in-the-loop} classification that combines dynamic arguments with legal case-based reasoning to create a system that can elaborate its decisions and adapt to new situations.
In this system, human analysts can add new factors to update system results and make recommendations to the classification algorithm as supervisors.}

\REO{\p{Across that span of time}, with the further development of Internet Technology, security systems become an urgent need.
However, the use of \REO{human-in-the-loop} is not limited to virus detection and fraudulent information filtering and has also shown remarkable performance in issues such as privacy protection, authentication attack prevention, and spam filtering.}

\REO{\subsection{Code Production Tools}}

\begin{figure}[t]
    \centering
    \includegraphics[width=0.88\linewidth]{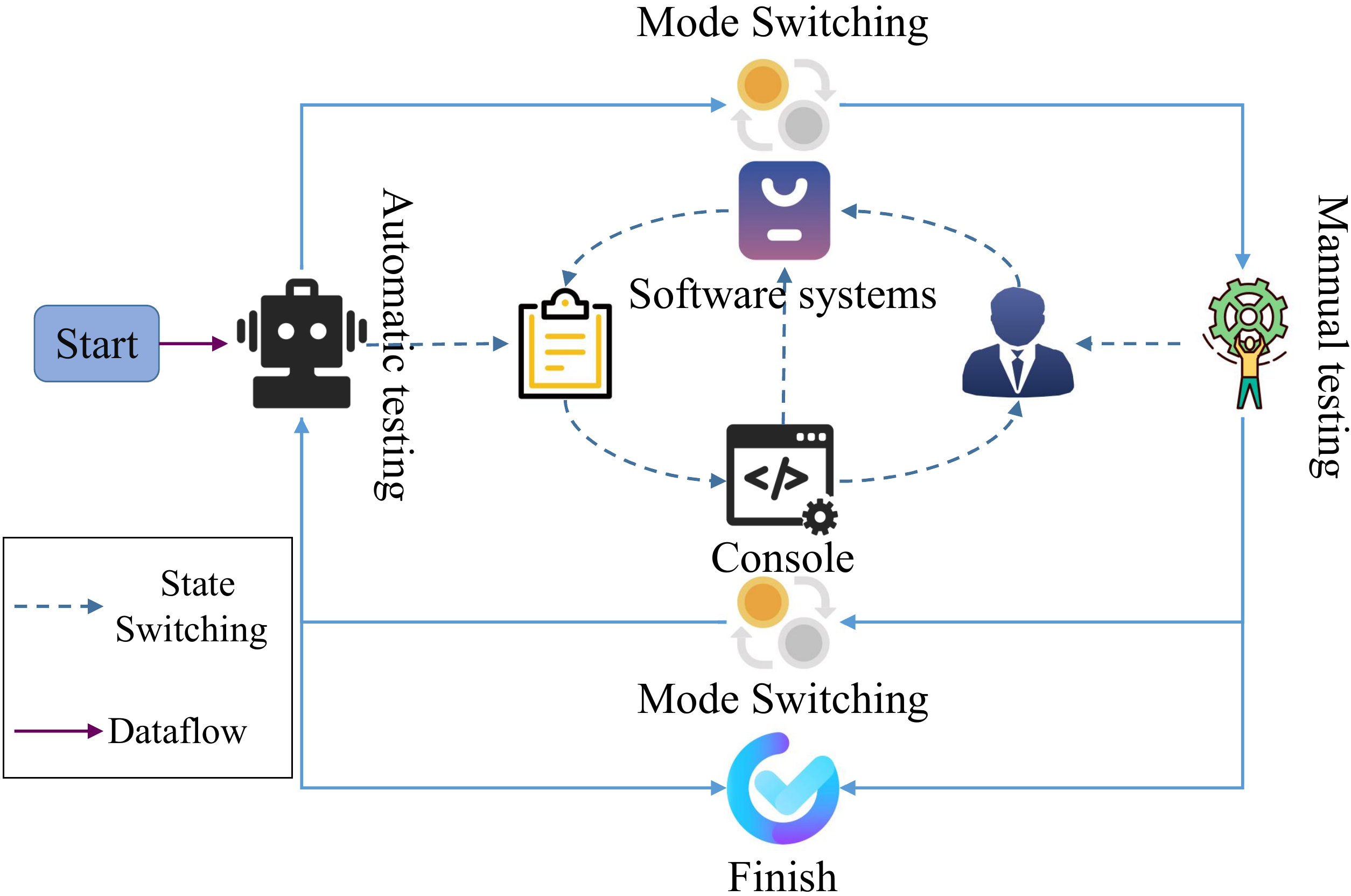}
    \caption{With the help of \REO{human-in-the-loop} system, human can collaborate a project with computer.}
    \label{fig_syx_3}
\end{figure}

\REO{
The programming of programs and the training of models are \p{essential subtasks} of artificial intelligence, and this two tasks are highly coupled with humans. However, with the establishment of the open-source community, \p{various} codes and model resources have been disclosed, \p{making simple program writing and model training a repetitive work and thus leading to an automated basis for the model.} The \p{fundamental} difficulty is \p{making} integration tools learn to splice existing components, which requires integration tools to learn human integration experience. Under the guidance of this idea, there have been some tools for automation. With the help of these \p{practical tools}, developers can collaborate on existing projects, rather than writing projects from scratch (Figure~\ref{fig_syx_3}).}

\REO{
Software testing was first addressed. MacHiry~\etal~\cite{machiry2013dynodroid} \p{designed} a test generation system called Dynodroid for fuzzing unmodified Android applications.
Dynodroid decomposed an application into a set of event-driven programs.
Dynodroid \p{employed} the Android framework to generate a series of events and automatically executed these generated events to interact with the surrounding environment. It collected interactive feedback to program functions. The most worth mentioning is that \p{Dynodroid would use manual intervention mode when necessary, and also tried to exploit input feedback to adjust the frame to generate new output dynamically.}
Through the above approach, Dynodroid implemented fuzz testing of unmodified Android applications in a human-machine \p{hybrid manner}.
Yan~\etal~\cite{shoshitaishvili2017rise} \p{put forward} a human\& tool-centric vulnerability analysis system that leveraged humans (with varying levels of expertise) to perform well-defined subtasks. \p{The vulnerability analysis system had a larger scale of available programs and higher labor utilization than previous work.}
Besides, in the practical application situation, the coders pay attention to the test of the code and want to see the application that can make the program correction.
Bohme~\etal~\cite{bohme2020human} proposed the first \REO{human-in-the-loop} semi-automatic program repair framework called LEARN2FIX.
LEARN2FIX repaired bugs by negotiating with users to observe errors, trained an automatic error oracle through the bug samples marked in the negotiation and finally realized the automatic repair of the program. LEARN2FIX learned a sufficiently accurate automatic oracle with \p{relatively}  low labeling effort (e.g. 20 queries).
Productization tools of \REO{human-in-the-loop} systems in software engineering bring great convenience to programmers. In the future, the application of \REO{human-in-the-loop} in this field will expand from debugging and software testing to most coding programs. In particular, with the introduction of pre-trained models such as BERT~\cite{devlin2018bert} and GPT~\cite{radford2018improving}, human-machine collaborative programming is becoming a new focus.}

\begin{figure}[t]
    \centering
    \includegraphics[width=0.88\linewidth]{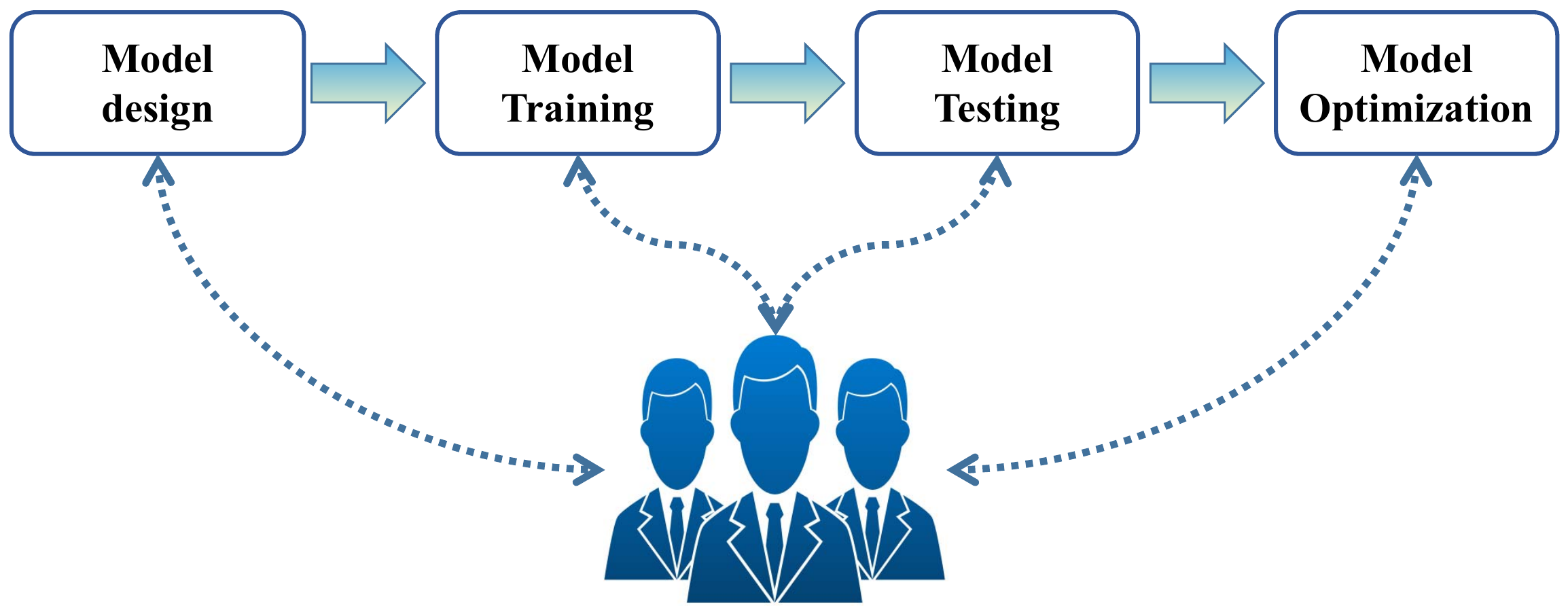}
    \caption{\REO{human-in-the-loop} system in machine learning algorithm optimizing.}
    \label{fig_syx_4}
\end{figure}

\REO{
With the development of machine learning, the training of deep models has become another \p{prevalent} task.
With the continuous exploration of model training, more researchers focus on constructing semi-automatic or even fully automatic training tools.
\p{The researchers make efforts to incorporate human knowledge during various stages of model training} (Fig.~\ref{fig_syx_4}).
To solve the problem that the underlying structure of the model is unknown in the model design process, Salam~\etal~\cite{salam2019human} \p{applied} a semi-automated \REO{human-in-the-loop} attribute design framework to \p{assist human analysts in converting} original attributes into classifications of valid derived properties of the question. The framework first provided a human analyst with k buckets containing promising original attribute selections based on a random walk-based heuristic. \p{It then iteratively allowed human analysts to deal with attributes involving attributes based on a scalable and efficient greedy heuristic algorithm.} The top-l derived attribute finally got a designed model frame, and the designer completed the detailed design interactively.
To address the model training process, the ultimate goal of an ML system becomes the problem of reducing the time it takes to get a deployable model from scratch.
MA~\etal~\cite{ma2018towards} \p{exploited} a human-machine collaboration system called Helix which optimized execution across iterations by appropriately reusing or recomputing intermediate results. It is worth pointing out that Helix \p{contained} a workflow management module as well as a visualization module to interact with people, thus enhancing the usability of the tool.
To solve the problem that transparency and operability are ignored in the process of model optimization,
Renner~\etal\cite{renner2020designing} focused on both transparency (how the system is explained to humans) and operability (how users provide feedback or coach the system) in interactive machine learning. It combined novel topic visualization techniques with a human-centered interactive topic modeling system. \p{Furthermore}, it revealed how users understand and interact with machine learning models and guided the further development of \REO{human-in-the-loop} systems.
}

\REO{\subsection{Simulation System}}

The simulation system \p{generally utilizes} virtual systems to simulate an object or workflow, and is widely applied in the decision-making of system construction, process forecast, and safety control. Due to the feature of these applications, interaction with a human is indispensable.
\REO{
\p{A good many researchers have been working towards the goal of improving the usability of the \REO{human-in-the-loop} in predicting and optimizing the integration of soldier systems.} Davidson~\etal~\cite{davidson2021improving} \p{presented} a semi-structured interview method with key informants, and in the review method, eight critical requirements for improving \REO{human-in-the-loop} simulation systems were identified. Addressing essential requirements can improve the ability of current \REO{human-in-the-loop} simulation tools to accommodate the military's need for human consideration early in the design process. In addition, the system can also be used for other planning problems such as cargo, transportation problems, and traffic management problems.
Other researchers are concerned with the simulation of processes, the most representative of integrating human-machine dialogue into a system.
Demirel~\etal~\cite{demirel2020digital} proposed a novel human-computer interaction method to evaluate the safety and performance of human-product interactions. \p{This method modeled humans into a computationally designed environment through digital human modeling technology.} With this system, the form and function of the characters in the workflow could be pre-designed, and ergonomic evaluation indicators could be used to implement the work cycle.
In terms of program control, Metzner~\etal~\cite{metzner2020system} proposed a simulated human-robot collaboration method that combined virtual reality, motion tracking, and standard simulation software for industrial robots. It used a virtual reality system to simulate the workplace of a robot and its collaborators to design a motion tracking system. The system captured human movements by fusing two above-mentioned subjects. Among other things, this approach would evaluate the performance of the human-robot collaboration system safety controls and the fulfillment of the defined requirements.}

\REO{\subsection{Search Engine}}

\begin{figure}[t]
    \centering
    \includegraphics[width=0.88\linewidth]{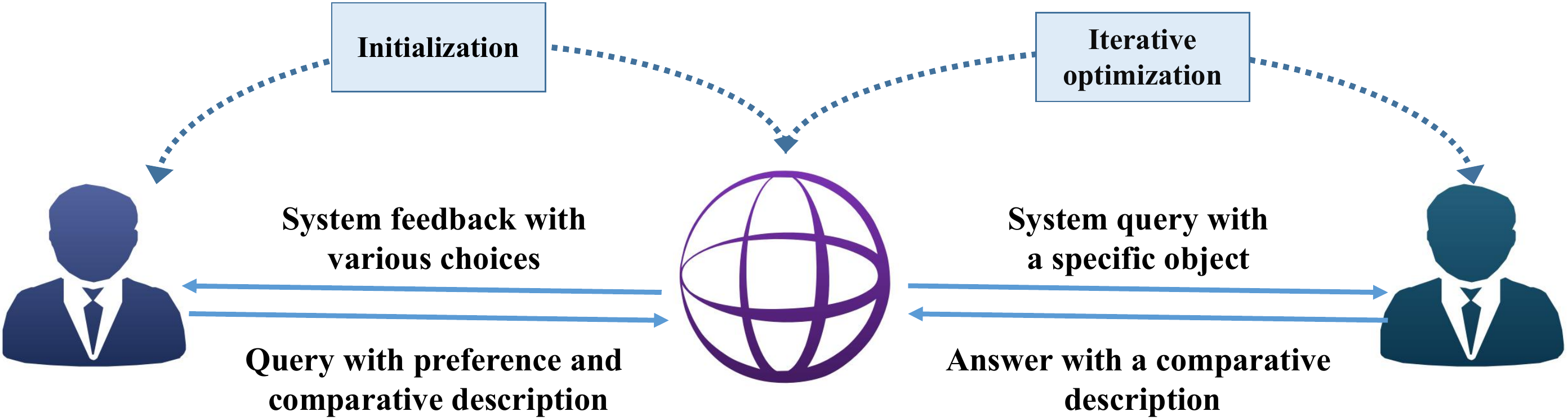}
    \caption{The \REO{human-in-the-loop} usage for image search system.}
    \label{fig_syx_5}
\end{figure}

\REO{
The automation and intelligence of search engines are also important research areas. We \p{discover that current search engine is still positioned as an human assistant, but if the search engine can further model the human work, it can be more practical.
At present, the researches on search engines are mainly divided into recommendation system and image search.}
Polisetty~\etal~\cite{polisetty2020information} \p{introduced} a new joint recommender system that is updated by continuously learning from user feedback. Recommender systems better prepare information from the three main research directions of information aggregation, enrichment, and recommendation by placing humans in the loop. The system features review summarization and rating prediction, with a web-based interface for improving methods through human-computer interaction.
With the development of computer vision, searching through images has become an important task \p{and the image search task has been well solved due to the fast development of deep learning.} However, the visual-text semantic matching task remains a significant challenge that affects interaction performance and search results. Since humans have \p{an inherent} advantage in handling image search tasks, trying to introduce users into the image search loop (Figure ~\ref{fig_syx_5}) has \p{excellent potential.}
Kovashka~\etal~\cite{kovashka2015whittlesearch} devised a new feedback model for image search that allowed users to interact, communicated preferences through visual comparisons, and ultimately determined user searches.
First, the user initiates a query; then the system initiates feedback on the user's query, and the user provides feedback; then the system collects user feedback and responds until the system no longer receives user feedback; finally, the system learns from the user feedback and updates the system. Compared to traditional passive and active methods, the system can provide more accurate search results with less user interaction through an iterative learning process.
Building on Kovashka~\etal, Plummer~\etal~\cite{plummer2019give} \p{put forward} an attribute-based interactive image search method, which can iteratively refine image search results using \REO{human-in-the-loop} feedback. Unlike before, the system trains a deep reinforcement model to learn which images are informative, rather than relying on manual measurements commonly exploited in previous work. Besides, this work extends conditional similarity networks to incorporate global similarity into training visual embeddings, resulting in more natural transitions when users explore learned similarity embeddings.
The system is self-updating to provide more accurate image search results.
With the help of \REO{human-in-the-loop}, we believe that search engines can continuously adapt to different changes through interactive learning to \p{meet searchers' personalized needs better.}}

\REO{
In addition to the above scenarios, \REO{human-in-the-loop} systems are also used in other fields ~\cite{wrede2019smart, zhu2020easierpath}, such as bioinformatics, supervisory healthcare, and crowd-sourcing work. For example, Fredrik Wrede~\etal~\cite{wrede2019smart} \p{utilized} \REO{human-in-the-loop} semi-supervised learning for stochastic gene regulation.
Rosenberg~\etal~\cite{rosenberg2016artificial} \p{applied} an artificial swarm intelligence scheduling system for crowd-sourcing tasks,
Li~\etal ~\cite{li2020explanations} \p{employed} a model checking method for \REO{human-in-the-loop} system.
With the further development of the systems based on human-computer interaction and the expansion of application scenarios, human-computer interaction will be applied on more occasions.}

\section{Discussion and Future Directions}
\label{sec:DF}

\REO{In this section, we first discuss some existing challenges and key issues of \ZJH{human-in-the-loop} for machine learning.
Next, we show future directions on CV, NLP, and applications based on \ZJH{human-in-the-loop.}}

\REO{\subsection{The Challenges and Discussion}}
\noindent
\textbf{How to add human experience and knowledge to computer vision tasks? }
By reviewing the previous works, we find that most \REO{human-in-the-loop} \ZJH{researches only focus} on natural language processing. Analyzing the reasons, \p{it is not} easy to directly allow people to interact with images effectively( except for direct labeling ), and add human experience and knowledge to the model throughout the cycle.
With the development of multi-modal technology, \p{utilizing} multi-modality for image representation \ZJH{can} be an effective way~\cite{wiriyathammabhum2016computer}.
\REO{It is particularly worth pointing out that Holzinger~\etal~\cite{holzinger2021towards} \p{adopted} Graph Neural Networks as a method of choice, enabling information fusion for multi-modal capability, \ZJH{which is a milestone.}}
\p{Besides, applying inverse reinforcement learning also seems to be a feasible and practical solution~\cite{arora2021survey}.}

\noindent
\textbf{How does the model learn human knowledge and experience from a higher dimension? }
The goal of \REO{human-in-the-loop} is to connect humans to the model loop \p{in a specific way}, so that the machine can learn human knowledge and experience during the loop. Most current methods achieve this goal through human data annotation \ZJH{which is only the most basic realization process.} \p{As the saying goes, it is better to teach a man fishing than to give him fish. Researchers are supposed to take how to help agents acquire this knowledge effectively into account~\cite{doan2017human}.}
Language is an experience accumulated in the human learning process. \p{Researchers currently focus on employing human intervention in dialogue to enable machines to learn human knowledge and intelligence from dialogue procedures iteratively~\cite{li2016dialogue}.}
In addition, numerous reasoning tasks contain higher-dimensional knowledge. By integrating humans into the reasoning loop, the machines can also learn more about human experience~\cite{cranor2008framework}.
Image quality evaluation and design tasks are a higher level of human activity. Although human aesthetics and design inspiration can constitute the theory\ZJH{; however,} more inspiration and aesthetics still come from human experience~\cite{amirpourazarian2021quality}. \ZJH{\p{If we find a productive way to allow the model to learn more expert experience, the model's improvement can be dramatic.}}

\noindent
\textbf{How to select key samples? }
The critical technology for the \REO{human-in-the-loop} is obtaining essential samples and labeling them with human intervention. At present, researchers mainly exploit confidence-based methods to obtain critical samples. This method plays an irreplaceable role in classification tasks~\cite{wan2020human,yang2020optimal}.
However, for other tasks \ZJH{(e.g. semantic segmentation, regression, and target detection tasks)}, \p{confidence is the evaluation that is not so noticeable and thus cannot reflect the improvement of the system.}
Active learning aims to train an accurate prediction model with the least cost by marking the examples that provide the most information. There are multiple mature and worthy reference methods in selecting criteria, and perhaps researchers can obtain inspiration from these methods~\cite{fu2013survey}.

\noindent
\textbf{How to construct an evaluation benchmark? }
To develop the entire community, \ZJH{providing an effective test benchmark is important.} At present, there is no uniform standard for \REO{human-in-the-loop} research benchmarks. To better explore this research topic, it is essential to study how to develop evaluation methods and benchmarks for \REO{human-in-the-loop} systems. Moreover, the formation of a unified benchmark is also conducive to the further refinement of research~\cite{chai2020human}. The current \REO{human-in-the-loop}-based research is a more influential direction for exploring ways that are more conducive to \REO{human-in-the-loop}. \ZJH{In addition to} creating the standards for these interaction methods, restricting and theorization are also particularly important.

\noindent
\REO{
\textbf{How to implement a general multitasking framework by \REO{human-in-the-loop}? }}
\ZJH{The real-world task is complex and in its current form, so it is not easy to completely solve it with one characterization~\cite{zhang2017human}.}
With the emergence of a unified large-scale pre-training model~\cite{tay2020efficient,khan2021transformers}, we have seen the hope of achieving a universal model through \REO{human-in-the-loop} fine-tuning. In particular, the current machine learning models are not as intelligent as humans, so it may be the next direction to consider using a suitable way to introduce human knowledge into large models.

\REO{\subsection{Future Directions}}
\REO{To facilitate more researchers developing more advanced Human-in-the-loop systems, we summarize the following concrete future directions for Human-in-the-loop NLP, Human-in-the-loop  CV, and  Human-in-the-loop  Real-world Applications.}

\REO{\subsubsection{Future Directions for Human-in-the-loop NLP Systems}}
\begin{itemize}
\setlength{\itemsep}{0pt}
\setlength{\parsep}{0pt}
\setlength{\parskip}{0pt}
\item
For systems like a chatbot, automatic summarization tool, or commercial machine translation, \ZJH{when interacting} with them, individuals can only give a reward signal to the one output that is sent to them, which leads to the sparsity in feedback concerning the size of the output space~\cite{kreutzer2020learning}.

\item
\REO{
In syntactic parsing tasks, exploring more intelligent questions about other types of parsing uncertainties based on human-in-the-loop and scaling the method to large unlabelled corpora or other languages \ZJH{are} important~\cite{he2016human}.}

\item
\REO{
In user-centered design and evaluation of a \REO{human-in-the-loop} topic modeling task, the key is considering trust or confidence~\cite{smith2018closing}.}

\item
\ZJH{In terms of AI safety,} among existing \ZJH{human-in-the-loop} techniques, some of them also allow malicious individuals to efficiently train models that serve their purpose, which may cause damage to all aspects of society. For example, they could exploit human feedback to fine-tune a language model to be more persuasive and manipulate humans' beliefs, \p{instill radical ideas, commit fraud, etc.~\cite{stiennon2020learning}.}
\end{itemize}

\REO{\subsubsection{Future Directions for Human-in-the-loop CV Systems}}
\begin{itemize}
\setlength{\itemsep}{0pt}
\setlength{\parsep}{0pt}
\setlength{\parskip}{0pt}
\item
\REO{
In the image restoration task, it is necessary to pay attention to predictive parameter optimization based on supervised regression models and scientifically analyzing correlations between the parameters of different algorithms based on the human-in-the-loop methods~\cite{roels2019human}.}

\item
\REO{
In image enhancement tasks, using active learning to help users obtain a better estimation of cluster membership with the fewest image enhancements of images is important~\cite{kapoor2014collaborative}.}
\end{itemize}

\REO{\subsubsection{Future Directions both NLP and CV Systems}}
\begin{itemize}
\setlength{\itemsep}{0pt}
\setlength{\parsep}{0pt}
\setlength{\parskip}{0pt}
\item
Human supervision may be preferable due to various levels of expertise and with the increase of work overload, erroneous decisions are potential to occur~\cite{jwo2021smart}.

\item
Collect and share more human feedback datasets for different tasks of NLP and CV~\cite{chai2020human}.

\item
We should consider user credibility to affect the influence of their annotations by analyzing the quality of provided feedback\cite{karmakharm2019journalist}.

\item
More rigorous in-depth user studies need to be designed and conducted to evaluate the effectiveness and robustness of \REO{human-in-the-loop} frameworks in addition to model performance~\cite{smith2018closing}.

\item
\REO{For generative tasks, an explicit function can be defined by user feedback to evaluate and collect the generated signals~\cite{kreutzer2020learning}.}

\item
\REO{It is vital \ZJH{to find} an efficient way to dynamically pick up the most representative and valuable feedback to collect~\cite{settles2011closing}.}

\item
\REO{It is crucial to have a more friendly manner of showing what the model has learned from feedback and what kind of feedback. Specifically, it is possible to explore the process of changing the model in an attempt to visualize the results of the manual feedback~\cite{lee2017human}.}

\end{itemize}

\REO{\subsubsection{Future Directions in Real-world Applications}}
\begin{itemize}
\setlength{\itemsep}{0pt}
\setlength{\parsep}{0pt}
\setlength{\parskip}{0pt}
\item
\REO{It is fundamental to choose an appropriate artificial intervention time, especially the tasks with solid demand for reliability and safety~\cite{marquand2021automated}.}

\item
\REO{For a system with human-computer interaction, users' expectations of experience usually take precedence over performance~\cite{dudley2018review,shilton2018values}.}
\item
\REO{Modeling the sensor signal and solving the unified coding of abstract and concrete information is an essential problem in the process of human-computer interaction~\cite{jolfaei2022guest}.}
\item
\REO{Human intervention remains on superficial judgments (such as acceptance/rejection or direction), and exploring more complex feedback is also a critical issue in \ZJH{human-in-the-loop} applications~\cite{xu2022transitioning}.}
\item
\REO{Human-in-the-loop-based systems should have high robustness and generalization ability to domain changes, disturbances, and ``out-of-range'' samples~\cite{zhou2021domain}.}
\end{itemize}

\section{Conclusion}
\label{sec:conclusion}

In this paper, we review existing studies in \REO{human-in-the-loop} techniques for machine learning.
We first discuss the work of improving model performance from data processing.
\REO{\ZJH{We divide this section} into three parts: Data Preprocessing, Data Annotation, and Iterative Labeling according to the \REO{human-in-the-loop} data processing pipeline. The core question is how to achieve more significant performance with fewer samples from a data perspective. Through the investigation of \REO{human-in-the-loop}-based data processing methods, we \ZJH{find} that the current \REO{human-in-the-loop}-based data processing methods focus more on how to use semi-supervised methods for data collection and \ZJH{annotation} than on how to select and \ZJH{identify} key samples in the dataset.}
Then, we discuss the work of improving model performance through interventional model training.
\REO{
In this part, we divide into sub-module focusing on NLP and CV according to the target task. The core issue is how to promote the breakthrough of the model to crucial problems by adding human high-dimensional knowledge to the model. By investigating the work of \REO{human-in-the-loop} in model training, we found that most \REO{human-in-the-loop} exists in model training by participating in simple data increments. Exploring model training methods that effectively integrate human knowledge is the core of solving such problems. Methods such as inverse reinforcement learning and multimodal learning seem to be effective solutions.
}
Finally, we discuss the design of the system independent ``\REO{human-in-the-loop}".
\REO{
In this section, we provide an overview of several \REO{human-in-the-loop} real-world applications tasks, and we divide the content into Security Systems, Code Production Tools, Simulation Systems, and Search Engines by task type. Through the investigation of \REO{human-in-the-loop} applications, we \ZJH{find} that the coordination of considerable variables in practical applications is more complicated than a single variable in academic research. In actual production, although the system integrating \REO{human-in-the-loop} has advantages, how to quickly integrate high-level human knowledge in actual production and maintain the robustness of the system is a problem worthy of further consideration.
}
Besides, we provide open challenges and opportunities and introduce some exciting questions.
\REO{We first \ZJH{discuss} some challenges and problems of the current \REO{human-in-the-loop} in machine learning. On this basis, we further \ZJH{clarify} the future directions in terms of CV, NLP, and applications. The establishment of artificial intelligence models that can effectively integrate human high-dimensional knowledge has development potential, especially with the research of large-scale pre-training models, the human-machine hybrid method based on few-shot has broad prospects.}
\\

\noindent\textbf{Acknowledgment.}
This work was supported in part by the 2020 East China Normal University Outstanding Doctoral Students Academic Innovation Ability Improvement Project (YBNLTS2020-042), the Science and Technology Commission of Shanghai Municipality (19511120200).

\small{
\bibliographystyle{IEEEtran}
\bibliography{total}}

\end{document}